\documentclass[journal]{assets/IEEEtran}

\usepackage{cite}
\usepackage[flushleft]{threeparttable} 
\usepackage{graphicx}
\usepackage{textcomp}
\usepackage{tikz} 
\usetikzlibrary{positioning}
\usepackage{amsmath} 
\usepackage[ruled,linesnumbered]{algorithm2e} 
\usepackage{bm} 
\usepackage{url}
\usepackage{pgfplots} 
\usepgfplotslibrary{fillbetween} 
\usepackage{soul} 
\usepackage[flushleft]{threeparttable} 
\usepackage{enumitem} 
\usepackage{array} 
\usepackage{mathtools} 
\usepackage{filecontents}
\usepackage{xcolor} 
\makeatletter
\let\MYcaption\@makecaption
\makeatother
\usepackage[font=footnotesize]{subcaption}
\makeatletter
\let\@makecaption\MYcaption
\makeatother

\makeatletter 
\newcount\SOUL@minus
\makeatother  

\DeclareMathOperator*{\argmax}{arg\,max}

\newcommand{\calc}[2]{\pgfmathparse{#1}\pgfmathprintnumber[assume math mode=true,fixed zerofill,precision=#2]{\pgfmathresult}}

\definecolor{cross_eat}{RGB}{0,245,255}
\definecolor{cross_drink}{RGB}{255,0,230}
\definecolor{ctc_eat}{RGB}{0,0,255}
\definecolor{ctc_drink}{RGB}{255,0,0}

\newcommand{\fTOREBAIneThrGen}{0.274922}
\newcommand{\fTClemsonIneThrGen}{0.362125} 

\newcommand{\fTOREBAIneTwoSOTAGen}{0.778} 
\newcommand{\fTOREBAVidTwoSOTAGen}{0.853} 

\newcommand{\fTOREBAIneTwoGen}{0.830752} 
\newcommand{\fTOREBAIneTwoEat}{0.798450} 
\newcommand{\fTOREBAIneTwoDri}{0.637500} 
\newcommand{\fTOREBAIneTwoEatDri}{0.783372} 
\newcommand{\fTOREBAVidTwoGen}{0.857939} 
\newcommand{\fTOREBAVidTwoEat}{0.841495} 
\newcommand{\fTOREBAVidTwoDri}{0.858896} 
\newcommand{\fTOREBAVidTwoEatDri}{0.843149} 
\newcommand{\fTClemsonIneTwoGen}{0.781242} 
\newcommand{\fTClemsonIneTwoEat}{0.742684} 
\newcommand{\fTClemsonIneTwoDri}{0.733333} 
\newcommand{\fTClemsonIneTwoEatDri}{0.741431} 
\newcommand{\fTOREBAIneKyrTwoGen}{0.740292} 
\newcommand{\fTOREBAIneKyrTwoEat}{0.732377} 
\newcommand{\fTOREBAIneKyrTwoDri}{0.657143} 
\newcommand{\fTOREBAIneKyrTwoEatDri}{0.726334} 
\newcommand{\fTClemsonIneKyrTwoGen}{0.727623} 
\newcommand{\fTClemsonIneKyrTwoEat}{0.672683} 
\newcommand{\fTClemsonIneKyrTwoDri}{0.641208} 
\newcommand{\fTClemsonIneKyrTwoEatDri}{0.668483} 
\newcommand{\fTOREBAIneHeyTwoGen}{0.799097} 
\newcommand{\fTOREBAIneHeyTwoEat}{0.772304} 
\newcommand{\fTOREBAIneHeyTwoDri}{0.695652} 
\newcommand{\fTOREBAIneHeyTwoEatDri}{0.764883} 
\newcommand{\fTClemsonIneHeyTwoGen}{0.782821} 
\newcommand{\fTClemsonIneHeyTwoEat}{0.680153} 
\newcommand{\fTClemsonIneHeyTwoDri}{0.696545} 
\newcommand{\fTClemsonIneHeyTwoEatDri}{0.682542} 
\newcommand{\fTOREBAVidSloTwoGen}{0.793085} 
\newcommand{\fTOREBAVidSloTwoEat}{0.750754} 
\newcommand{\fTOREBAVidSloTwoDri}{0.565854} 
\newcommand{\fTOREBAVidSloTwoEatDri}{0.730397} 

\newcommand{\avgFVSinGre}{0.839850}
\newcommand{\fVOREBAIneSinGen}{0.920043} 
\newcommand{\fVOREBAIneSinEatDri}{0.899946} 
\newcommand{\fVOREBAVidSinGen}{0.851707} 
\newcommand{\fVOREBAVidSinEatDri}{0.821330} 
\newcommand{\fVClemsonIneSinGen}{0.792051} 
\newcommand{\fVClemsonIneSinEatDri}{0.766814} 
\newcommand{\avgFTSinGre}{0.834108}
\newcommand{\fTOREBAIneSinGen}{0.855441} 
\newcommand{\fTOREBAIneSinEat}{0.837153} 
\newcommand{\fTOREBAIneSinDri}{0.770270} 
\newcommand{\fTOREBAIneSinEatDri}{0.831672} 
\newcommand{\fTOREBAVidSinGen}{0.875497} 
\newcommand{\fTOREBAVidSinEat}{0.868762} 
\newcommand{\fTOREBAVidSinDri}{0.761290} 
\newcommand{\fTOREBAVidSinEatDri}{0.859393} 
\newcommand{\fTClemsonIneSinGen}{0.808387} 
\newcommand{\fTClemsonIneSinEat}{0.773087} 
\newcommand{\fTClemsonIneSinDri}{0.863208} 
\newcommand{\fTClemsonIneSinEatDri}{0.782899} 

\DeclareRobustCommand{\relImpTOwnOREBAVidGen}{\calc{(\fTOREBAVidSinGen-\fTOREBAVidTwoGen)/\fTOREBAVidTwoGen*100}{1}}
\DeclareRobustCommand{\relImpTOwnOREBAIneEatDri}{\calc{(\fTOREBAIneSinEatDri-\fTOREBAIneTwoEatDri)/\fTOREBAIneTwoEatDri*100}{1}}
\DeclareRobustCommand{\relImpTOwnOREBAVidEatDri}{\calc{(\fTOREBAVidSinEatDri-\fTOREBAVidTwoEatDri)/\fTOREBAVidTwoEatDri*100}{1}}
\DeclareRobustCommand{\relImpTOwnClemsonIneGen}{\calc{(\fTClemsonIneSinGen-\fTClemsonIneTwoGen)/\fTClemsonIneTwoGen*100}{1}}

\DeclareRobustCommand{\relImpTSOTAOREBAVidGen}{\calc{(\fTOREBAVidSinGen-\fTOREBAVidSloTwoGen)/\fTOREBAVidSloTwoGen*100}{1}}
\DeclareRobustCommand{\relImpTSOTAOREBAIneEatDri}{\calc{(\fTOREBAIneSinEatDri-\fTOREBAIneHeyTwoEatDri)/\fTOREBAIneHeyTwoEatDri*100}{1}}
\DeclareRobustCommand{\relImpTSOTAOREBAVidEatDri}{\calc{(\fTOREBAVidSinEatDri-\fTOREBAVidSloTwoEatDri)/\fTOREBAVidSloTwoEatDri*100}{1}}
\DeclareRobustCommand{\relImpTSOTAClemsonIneGen}{\calc{(\fTClemsonIneSinGen-\fTClemsonIneHeyTwoGen)/\fTClemsonIneHeyTwoGen*100}{1}}

\DeclareRobustCommand{\impDetTOwnOREBAVidGen}{{\calc{\fTOREBAVidTwoGen}{3}}$\,\to\,${\calc{\fTOREBAVidSinGen}{3}}}
\DeclareRobustCommand{\impDetTOwnOREBAIneEatDri}{{\calc{\fTOREBAIneTwoEatDri}{3}}$\,\to\,${\calc{\fTOREBAIneSinEatDri}{3}}}
\DeclareRobustCommand{\impDetTOwnOREBAVidEatDri}{{\calc{\fTOREBAVidTwoEatDri}{3}}$\,\to\,${\calc{\fTOREBAVidSinEatDri}{3}}}
\DeclareRobustCommand{\impDetTOwnClemsonIneGen}{{\calc{\fTClemsonIneTwoGen}{3}}$\,\to\,${\calc{\fTClemsonIneSinGen}{3}}}

\DeclareRobustCommand{\impDetTSOTAOREBAVidGen}{{\calc{\fTOREBAVidSloTwoGen}{3}}$\,\to\,${\calc{\fTOREBAVidSinGen}{3}}}
\DeclareRobustCommand{\impDetTSOTAOREBAIneEatDri}{{\calc{\fTOREBAIneHeyTwoEatDri}{3}}$\,\to\,${\calc{\fTOREBAIneSinEatDri}{3}}}
\DeclareRobustCommand{\impDetTSOTAOREBAVidEatDri}{{\calc{\fTOREBAVidSloTwoEatDri}{3}}$\,\to\,${\calc{\fTOREBAVidSinEatDri}{3}}}
\DeclareRobustCommand{\impDetTSOTAClemsonIneGen}{{\calc{\fTClemsonIneHeyTwoGen}{3}}$\,\to\,${\calc{\fTClemsonIneSinGen}{3}}}

\begin{document}

\title{Single-stage intake gesture detection using \\ CTC loss and extended prefix beam search}
\author{Philipp~V.~Rouast,~\IEEEmembership{Member,~IEEE,}
        Marc~T.~P.~Adam%
\thanks{The authors are with the School of Electrical Engineering and Computing, The University of Newcastle, Callaghan, NSW 2308, Australia. E-mail: philipp.rouast@uon.edu.au, marc.adam@newcastle.edu.au.}%
}

\markboth{Journal of \LaTeX\ Class Files,~Vol.~14, No.~8, August~2020}%
{Shell \MakeLowercase{\textit{et al.}}: Bare Demo of IEEEtran.cls for IEEE Journals}

\maketitle

\begin{abstract}
Accurate detection of individual intake gestures is a key step towards automatic dietary monitoring.
Both inertial sensor data of wrist movements and video data depicting the upper body have been used for this purpose.
The most advanced approaches to date use a two-stage approach, in which (i) frame-level intake probabilities are learned from the sensor data using a deep neural network, and then (ii)
sparse intake events are detected by finding the maxima of the frame-level probabilities.
In this study, we propose a single-stage approach which directly decodes the probabilities learned from sensor data into sparse intake detections.
This is achieved by weakly supervised training using Connectionist Temporal Classification (CTC) loss, and decoding using a novel extended prefix beam search decoding algorithm.
Benefits of this approach include (i) end-to-end training for detections, (ii) simplified timing requirements for intake gesture labels, and (iii) improved detection performance compared to existing approaches.
Across two separate datasets, we achieve relative $F_1$ score improvements between \relImpTOwnOREBAVidEatDri\% and \relImpTOwnOREBAIneEatDri\% over the two-stage approach for intake detection and eating/drinking detection tasks, for both video and inertial sensors.
\end{abstract}



\section{Introduction}
\label{sec:introduction}

\IEEEPARstart{A}{ccurate} information on dietary intake forms the basis of assessing a person's diet and delivering dietary interventions.
To date, such information is typically sourced through memory recall or manual input, for example via dietitians \cite{block1982review} or smartphone apps used to log meals.
Such methods are known to require substantial time and manual effort, and are subject to human error \cite{lichtman1992discrepancy}.
Hence, recent research has investigated how dietary monitoring can be partially automated using sensor data and machine learning \cite{vu2017wearable}.

Detection of individual intake gestures in particular is a key step towards automatic dietary monitoring.
Wrist-worn inertial sensors provide an unobtrusive way to detect these gestures.
Early work on the Clemson dataset, established in 2012, used threshold values for detection from inertial data \cite{dong2012new}.
More recent developments include the use of machine learning to learn features automatically \cite{kyritsis2017food} and learning from video, which has become more practical with emerging spherical camera technology \cite{rouast2019learning}\cite{qiu2019assessing}.
Research on the OREBA dataset showed that frontal video data can exhibit even higher accuracies in detecting eating gestures than inertial data \cite{rouast2020oreba}.

The two-stage approach introduced by Kyritsis et al. \cite{kyritsis2020modeling} is currently the most advanced approach benchmarked on publicly available datasets for both inertial \cite{kyritsis2020modeling} and video data \cite{rouast2019learning}.
It first estimates frame-level intake probabilities using deep learning, which are then searched for maxima to detect intake events.
Thereby, the two-stage approach builds on a predefined gap between intake gestures in the second stage.

\DeclareRobustCommand{\relImpTOREBAIneSOTA}{\calc{(\fTOREBAIneSinGen-\fTOREBAIneTwoSOTAGen)/\fTOREBAIneTwoSOTAGen*100}{1}}
\DeclareRobustCommand{\relImpTOREBAVidSOTA}{\calc{(\fTOREBAVidSinGen-\fTOREBAVidTwoSOTAGen)/\fTOREBAVidTwoSOTAGen*100}{1}}

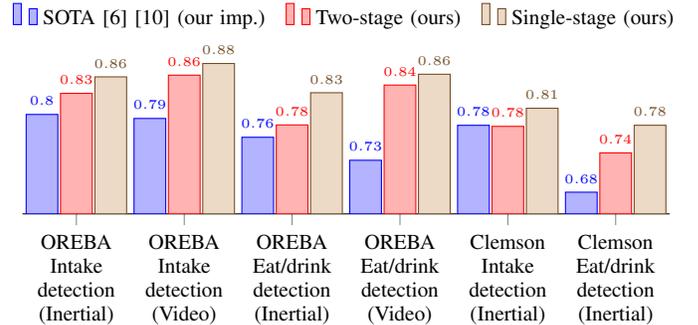
\begin{figure}
\begin{tikzpicture}
\begin{axis}[
    ybar=1,
    bar width=12,
    ymin=0.65, ymax=0.9,
    axis y line=none,
    axis x line*=bottom,
    legend style={at={(0.5,1.3)}, anchor=north,legend columns=-1, /tikz/every even column/.append style={column sep=0.2cm}, font=\footnotesize, legend style={draw=none}},
    symbolic x coords={ogeni,ogenv,oeati,oeatv,cgeni,ceati},
    xtick=data,
    xticklabels={OREBA\\Intake\\detection\\(Inertial),OREBA\\Intake\\detection\\(Video),OREBA\\Eat/drink\\detection\\(Inertial),OREBA\\Eat/drink\\detection\\(Video),Clemson\\Intake\\detection\\(Inertial),Clemson\\Eat/drink\\detection\\(Inertial)},
    xticklabel style={align=center},
    nodes near coords,
    every node near coord/.append style={font=\tiny},
    nodes near coords align={vertical},
    ticklabel style = {font=\footnotesize},
    width=1.15\columnwidth, height=3.8cm]
\addplot coordinates {(ogeni,\fTOREBAIneHeyTwoGen) (ogenv,\fTOREBAVidSloTwoGen) (oeati,\fTOREBAIneHeyTwoEatDri) (oeatv,\fTOREBAVidSloTwoEatDri) (cgeni,\fTClemsonIneHeyTwoGen) (ceati,\fTClemsonIneHeyTwoEatDri)};
\addplot coordinates {(ogeni,\fTOREBAIneTwoGen) (ogenv,\fTOREBAVidTwoGen) (oeati,\fTOREBAIneTwoEatDri) (oeatv,\fTOREBAVidTwoEatDri) (cgeni,\fTClemsonIneTwoGen) (ceati,\fTClemsonIneTwoEatDri)};
\addplot coordinates {(ogeni,\fTOREBAIneSinGen) (ogenv,\fTOREBAVidSinGen) (oeati,\fTOREBAIneSinEatDri) (oeatv,\fTOREBAVidSinEatDri) (cgeni,\fTClemsonIneSinGen) (ceati,\fTClemsonIneSinEatDri)};
\legend{SOTA \cite{rouast2019learning}\cite{heydarian2020deep} (our imp.),Two-stage (ours),Single-stage (ours)}
\end{axis}
\end{tikzpicture}
\caption{$F_1$ scores for our two-stage and single-stage models in comparison with the state of the art (SOTA). Our single-stage models see relative improvements between \relImpTSOTAClemsonIneGen\% and \relImpTSOTAOREBAVidEatDri\% over our implementations of the SOTA for inertial \cite{heydarian2020deep} and video modalities \cite{rouast2019learning}, and relative improvements between \relImpTOwnOREBAVidEatDri\% and \relImpTOwnOREBAIneEatDri\% over our own two-stage models for intake detection and eating/drinking detection across the OREBA and Clemson datasets.}
\label{fig:sota}
\end{figure}

\begin{figure*}[ht]
\centering
\includegraphics[width=\textwidth]{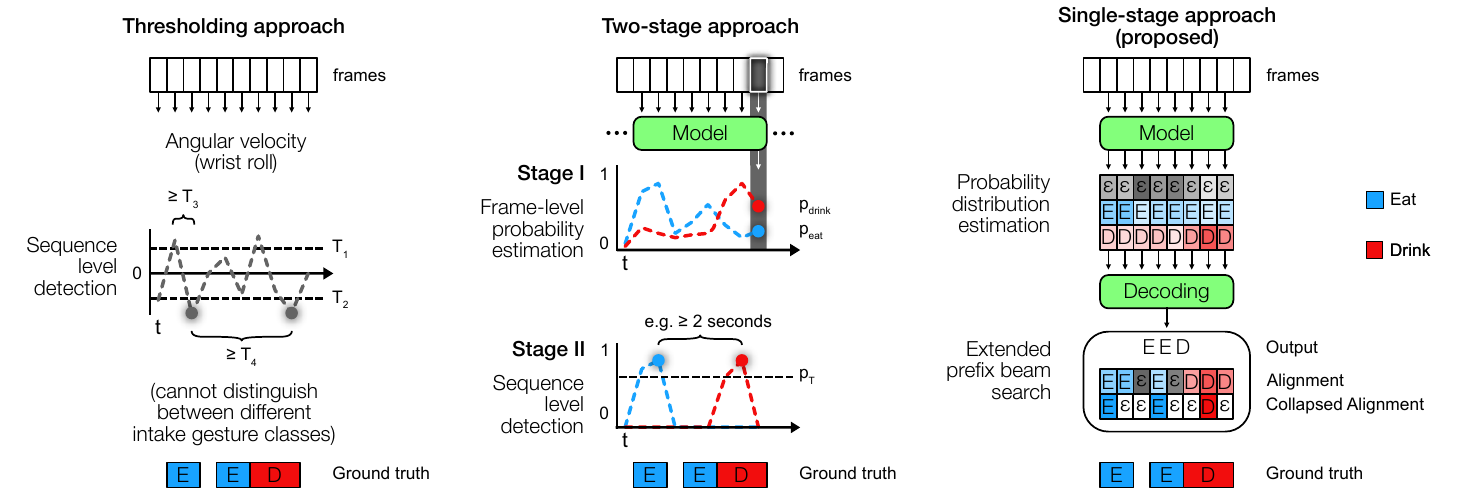}
\caption{Comparing existing approaches (left, center) to the proposed approach (right): The thresholding approach \cite{dong2012new} (left) searches the angular velocity for values that breach the thresholds $T_1$ and $T_2$. The two-stage approach \cite{kyritsis2020modeling} (center) independently estimates frame-level probabilities, which are then searched for maxima on the video level (generalized to two gesture classes here). The proposed single-stage approach (right) directly decodes the estimated probability distribution $p(c|x_t)$ using extended prefix beam search, after which token sequences in the most probable alignment $\hat{A}$ are collapsed to yield the result.}
\label{fig:stages}
\end{figure*}

In this paper, we propose a single-stage approach which directly decodes the probabilities learned from sensor data into sparse intake event detections.
We achieve this by weakly supervised training \cite{huang2016connectionist} of the underlying deep neural network with Connectionist Temporal Classification (CTC) loss, and decoding the probabilities using a novel extended prefix beam search algorithm.
Compared to the existing approaches in the literature, our study makes four key contributions:

\begin{enumerate}
	\item \textbf{Single-stage approach.} This is the first study that applies a single-stage approach allowing for end-to-end training with a loss function that directly addresses the intake gesture detection task. Thereby, we avoid the predefined gap between subsequent intake gestures in the second stage of two-stage models \cite{kyritsis2020modeling}\cite{rouast2019learning}.
	\item \textbf{Simplified labels.} The proposed approach requires information about occurrence and order of intake gestures, but not their exact timing. Hence, it is particularly suitable for intake gestures, whose start and end times are fuzzy in nature and time-consuming to determine.
	\item \textbf{Improved performance.} Our single-stage models outperform two-stage models on the OREBA and Clemson datasets, including the current state of the art (SOTA) \cite{rouast2019learning} \cite{heydarian2020deep} and two-stage versions of our models, see Fig. \ref{fig:sota}.
	\item \textbf{Intake gesture detection.} This is the first study to perform simultaneous localization and classification\footnote{For the purpose of this study, gesture \textit{detection} refers to temporal localization and simultaneous classification of a gesture (e.g., as a generic intake gesture, or as an eating or drinking gesture).} of intake gestures. While we use the example of eating and drinking, the approach could also be applied to more fine-grained analysis of dietary intake given appropriate data.
\end{enumerate}

The remainder of the paper is organized as follows:
In Section \ref{sec:related}, we discuss the related literature on CTC and intake gesture detection.
Our proposed method is introduced in Section \ref{sec:method}, including a complete pseudo-code listing of our proposed decoding algorithm.
We present and analyse the evaluation of our proposed model and the SOTA on two datasets in Section \ref{sec:experiments}.
Finally, we discuss the relative merits of the single-stage and two-stage approaches in Section \ref{sec:discussion} and conclude in Section \ref{sec:conclusion}.


\section{Related Research}
\label{sec:related}

\subsection{Intake gesture detection}

Intake gesture detection involves the determination of the timestamps at which a person moved their hands to ingest food or drink during an eating occasion.
It is one of the three elements of automatic dietary monitoring, which also encompasses classification of the consumed type of food, and estimation of the consumed quantity of food.
Sensors that carry a signal appropriate for the detection of intake gestures include inertial sensors mounted to the wrist \cite{heydarian2019assessing} and video recordings \cite{rouast2019learning}.
Note that information on eating events can also be derived from chewing and swallowing monitored using audio \cite{amft2005analysis} \cite{amft2009bite}, electromyography \cite{zhang2018monitoring}\cite{zhang2020retrieval}, and piezoelectric sensors \cite{sazonov2012sensor}.
There are also other recent video-based approaches based on skeletal and mouth \cite{konstantinidis2020validation} as well as food, hand and face \cite{qiu2019assessing} features extracted using deep learning.
For inertial data, there is recent work on in-the-wild monitoring \cite{kyritsis2020data}.
In the following, we focus on two main approaches for inertial and video data that have been benchmarked on publicly available datasets:

\subsubsection{Thresholding approach}

In 2012, Dong et al. \cite{dong2012new} devised an easily interpretable thresholding approach which requires the angular velocity around the wrist to first surpass a positive threshold (e.g., rolling one way to pick up food), and then a negative threshold (e.g., rolling the other way to pass food to the mouth).
Refer to Fig. \ref{fig:stages} (left) for an illustration.
The approach selects these thresholds and two further parameters for minimum time amounts during and after a detection based on an exhaustive search of the parameter space.
Note that this approach is not generalizable to multiple gesture classes.

\subsubsection{Two-stage approach}

Kyritsis et al. \cite{kyritsis2020modeling} proposed a two-stage approach for detecting intake gestures from accelerometer and gyroscope data.
Rouast and Adam \cite{rouast2019learning} later adopted this approach for video data.
In this approach, the first stage produces frame-level estimates for the probability of intake versus non-intake.
These estimates are provided iteratively by a neural network trained on a sliding two-second context.
The second stage identifies the sparse video-level intake gesture timings by operating a thresholded maximum search on the frame-level estimates, constrained by a minimum distance of two seconds between detections.
Fig. \ref{fig:stages} (center) illustrates this approach generalized to two intake gesture classes.

While this approach is also relatively easy to interpret and works well in practice \cite{kyritsis2020data}, there are a few aspects that need to be considered.
Firstly, the second stage requires a predefined gap of two seconds between subsequent intake gestures.
This predefined gap implies that consecutive events occurring within two seconds of each other lead to false negatives.
Secondly, the loss function during neural network training is geared towards optimizing the frame-level predictions, not the video-level detections. 
In the present work, we propose an alternative approach by introducing a new single-stage training and decoding approach using CTC -- see Fig. \ref{fig:stages} (right).

\subsection{Connectionist temporal classification}

In 2006, Graves et al. \cite{graves2006connectionist} proposed connectionist temporal classification (CTC) to allow direct use of unsegmented input data in sequence learning tasks with recurrent neural networks (RNNs).
By interpreting network output as a probability distribution over all possible token sequences, they derived CTC loss, which can be used to train the network via backpropagation \cite{graves2008supervised}.
Hence, what sets CTC apart from previous approaches is the ability to label entire sequences, as opposed to producing labels independently in a frame-by-frame fashion.
 
While the original application of CTC was phoneme recognition \cite{graves2006connectionist}, researchers have applied it in various sequence learning tasks such as end-to-end speech recognition \cite{graves2014towards}, handwriting recognition \cite{liwicki2007novel}, and lipreading \cite{assael2016lipnet}.
Further, CTC has also been applied to sign language recognition from wrist-worn inertial sensor data \cite{dai2017sound}\cite{zhang2019myosign}.
In the most closely related prior research to the present work, Huang et al. \cite{huang2016connectionist} extended the CTC framework to enable weakly supervised learning of actions from video, simplifying the required labelling process.
To this day, CTC has neither been applied for temporal localization of actions from sensor data nor intake gesture detection.


\section{Proposed Method}
\label{sec:method}

Our proposed approach interprets the problem of intake gesture detection as a sequence labelling problem using CTC.
This allows us to operate within a \textit{single-stage} approach, meaning that inference is operationalized for a single time window of data, as exemplified in Fig. \ref{fig:example}:

\begin{itemize}
	\item A probability distribution over possible events for each time step is estimated using a neural network previously trained with \textit{CTC loss} \cite{graves2006connectionist}.
	\item These probabilities are decoded using \textit{extended prefix beam search} and collapsed to derive the gesture timings.
\end{itemize}

We start by introducing the concept of alignments as well as the CTC loss function.
Then, we describe greedy decoding and prefix beam search as alternative decoding algorithms which provide the motivation for our extension.
Finally, we introduce the proposed extended prefix beam search.

\subsection{Alignment between sensor data and labels}

In many pattern recognition tasks involving the mapping of input sequences $X$ to corresponding output sequences $Y$, we encounter challenges relating to the alignment between the elements of $X$ and $Y$.
This is because real-world sensor data cannot always be aligned with fixed-size tokens: In handwriting recognition, for example, some written letters in $X$ are spatially wider than others, unlike the fixed-size tokens in $Y$ \cite{liwicki2007novel}.
A similar challenge arises in intake gesture detection, where gesture events can have various durations.

To account for the dynamic size of events in the input, we create an alignment $A$ by using the token in question multiple times \cite{hannun2017sequence}, such as in the example in Fig. \ref{fig:example}.
In addition, we introduce the blank token $\epsilon$ to allow separation of multiple instances of the same event class, $A=[E,E,\epsilon,E,E,D,D,D]$ in the example.
We derive the token sequence $Y$ from an alignment $A$ by first collapsing repeated tokens and then removing the blank token.
Hence, the token sequence for the example is $Y=[E,E,D]$, which correctly reflects the ground truth label.
Any one collapsed output token sequence $Y$ can have many possible corresponding alignments $A$.

\subsection{CTC loss for probability distribution estimation}
\label{sec:method:sub:ctc}

Suppose we have an input sequence $X$ of length $T$, the corresponding output token sequence $Y$, and possible tokens $\Sigma$.
Our network is designed to express a probability estimate $p(c|x_t)$ for each token $c$ in $\Sigma$ given the sensor input $x_t$ at time $t$.
Fig \ref{fig:example} continues the previous example to show what the network output $p(c|x_t)$ might look like.
The objective of CTC loss is to minimize the negative log-likelihood of $p(Y|X)$, which is the probability that the network predicts $Y$ when presented with $X$.
This probability can be expressed in the form given in Equation \ref{eq:prob} \cite{hannun2017sequence}, building on the individual tokens $a_t$ in all valid alignments $A_{X,Y}$ between $X$ and $Y$.

\begin{equation}
	p(Y|X) = \sum_{A\in A_{X,Y}} \prod_{t=1}^{T}p(c=a_t|x_t)
	\label{eq:prob}
\end{equation}

To train our single-stage networks for intake gesture detection, we use an implementation of CTC loss included in TensorFlow \cite{tensorflow2020ctc}.
This training process can be characterized as weakly supervised, since it only requires the less restrictive collapsed labels $Y$ which do not include timing information besides occurrence and order of the tokens.
An implication of using CTC loss is that our networks learn to make predictions differently than when trained with cross-entropy loss, as we explore further in Section \ref{sec:experiments:sub:effect}.
It also implies that examples are required to regularly contain multiple intake gestures for the network to learn properly (e.g., two eating and one drinking gesture in Fig. \ref{fig:example}).

\begin{figure}
\centering
\begin{tikzpicture}[scale=0.6]
\draw (-1.5,1) + (0.5,0.5) node{\small time};
\draw (0,1) + (0.5,0.5) node{\small $t_1$};
\draw (1,1) + (0.5,0.5) node{\small $t_2$};
\draw (2,1) + (0.5,0.5) node{\small $t_3$};
\draw (3,1) + (0.5,0.5) node{\small $t_4$};
\draw (4,1) + (0.5,0.5) node{\small $t_5$};
\draw (5,1) + (0.5,0.5) node{\small $t_6$};
\draw (6,1) + (0.5,0.5) node{\small $t_7$};
\draw (7,1) + (0.5,0.5) node{\small $t_8$};
\draw[thin] (-6.2,1) -- (8,1);
\node[rotate=90] at (-5.5,-1) {\small Dataset};
\draw (-3.3,0) + (0, 0.3) node{\small Data};
\draw (-3.3,-1) + (0,-0.3) node{\small Label};
\draw (-1.5,-0) + (0.5,0.5) node{\small frames};
\draw (0,0.25) rectangle (1,0.75);
\draw (1,0.25) rectangle (2,0.75);
\draw (2,0.25) rectangle (3,0.75);
\draw (3,0.25) rectangle (4,0.75);
\draw (4,0.25) rectangle (5,0.75);
\draw (5,0.25) rectangle (6,0.75);
\draw (6,0.25) rectangle (7,0.75);
\draw (7,0.25) rectangle (8,0.75);
\draw (-1.5,-0.8) + (0.5,0.5) node{\small ground};
\draw (-1.5,-1.2) + (0.5,0.5) node{\small truth};
\draw (0,-0.75) rectangle (2,-0.25);
\draw (0,-1) + (1, 0.5) node{\small Eat};
\draw (3,-0.75) rectangle (5,-0.25);
\draw (3,-1) + (1, 0.5) node{\small Eat};
\draw (5,-0.75) rectangle (8,-0.25);
\draw (5,-1) + (1.5, 0.5) node{\small Drink};
\draw (-1.5,-2) + (0.5,0.5) node{\small $A_L$};
\draw (0,-2) + (0.5,0.5) node{\small E};
\draw (1,-2) + (0.5,0.5) node{\small E};
\draw (2,-2) + (0.5,0.5) node{\small $\epsilon$};
\draw (3,-2) + (0.5,0.5) node{\small E};
\draw (4,-2) + (0.5,0.5) node{\small E};
\draw (5,-2) + (0.5,0.5) node{\small D};
\draw (6,-2) + (0.5,0.5) node{\small D};
\draw (7,-2) + (0.5,0.5) node{\small D};
\draw (-1.5,-3) + (0.5,0.5) node{\small $Y_L$};
\draw (0.5,-3) + (0.5,0.5) node{\small E};
\draw (3.5,-3) + (0.5,0.5) node{\small E};
\draw (6,-3) + (0.5,0.5) node{\small D};
\draw[thin] (-6.2,-3) -- (8,-3);
\node[rotate=90] at (-5.5,-7) {\small Single-stage approach};
\draw (-3.3,-4) + (0,0) node{\small $p(c|x_t)$};
\draw (-1.5,-4) + (0.5,0.5) node{\small $\epsilon$};
\draw (0.2,-4.8) -- (0.05,-4.8) -- (0.05,-3.2) -- (0.2,-3.2);
\draw (7.8,-4.8) -- (7.95,-4.8) -- (7.95,-3.2) -- (7.8,-3.2);
\draw (0,-4) + (0.5,0.5) node{\footnotesize 0.3};
\draw (1,-4) + (0.5,0.5) node{\footnotesize 0.25};
\draw (2,-4) + (0.5,0.5) node{\footnotesize 0.6};
\draw (3,-4) + (0.5,0.5) node{\footnotesize 0.4};
\draw (4,-4) + (0.5,0.5) node{\footnotesize 0.5};
\draw (5,-4) + (0.5,0.5) node{\footnotesize 0.3};
\draw (6,-4) + (0.5,0.5) node{\footnotesize 0.1};
\draw (7,-4) + (0.5,0.5) node{\footnotesize 0.2};
\draw (-1.5,-4.5) + (0.5,0.5) node{\small E};
\draw (0,-4.5) + (0.5,0.5) node{\footnotesize 0.5};
\draw (1,-4.5) + (0.5,0.5) node{\footnotesize 0.6};
\draw (2,-4.5) + (0.5,0.5) node{\footnotesize 0.2};
\draw (3,-4.5) + (0.5,0.5) node{\footnotesize 0.35};
\draw (4,-4.5) + (0.5,0.5) node{\footnotesize 0.4};
\draw (5,-4.5) + (0.5,0.5) node{\footnotesize 0.3};
\draw (6,-4.5) + (0.5,0.5) node{\footnotesize 0.2};
\draw (7,-4.5) + (0.5,0.5) node{\footnotesize 0.3};
\draw (-1.5,-5) + (0.5,0.5) node{\small D};
\draw (0,-5) + (0.5,0.5) node{\footnotesize 0.2};
\draw (1,-5) + (0.5,0.5) node{\footnotesize 0.15};
\draw (2,-5) + (0.5,0.5) node{\footnotesize 0.2};
\draw (3,-5) + (0.5,0.5) node{\footnotesize 0.25};
\draw (4,-5) + (0.5,0.5) node{\footnotesize 0.1};
\draw (5,-5) + (0.5,0.5) node{\footnotesize 0.4};
\draw (6,-5) + (0.5,0.5) node{\footnotesize 0.7};
\draw (7,-5) + (0.5,0.5) node{\footnotesize 0.5};
\draw[thin] (-4.75,-5) -- (8,-5);
\draw (-3.3,-6) + (0, 0.3) node{\small Greedy};
\draw (-3.3,-6) + (0,-0.3) node{\small decoding};
\draw (-1.5,-6) + (0.5,0.5) node{\small $A_G$};
\draw (0,-6) + (0.5,0.5) node{\small E};
\draw (1,-6) + (0.5,0.5) node{\small E};
\draw (2,-6) + (0.5,0.5) node{\small $\epsilon$};
\draw (3,-6) + (0.5,0.5) node{\small $\epsilon$};
\draw (4,-6) + (0.5,0.5) node{\small $\epsilon$};
\draw (5,-6) + (0.5,0.5) node{\small D};
\draw (6,-6) + (0.5,0.5) node{\small D};
\draw (7,-6) + (0.5,0.5) node{\small D};
\draw (-1.5,-7) + (0.5,0.5) node{\small $Y_G$};
\draw (0.5,-7) + (0.5,0.5) node{\small E};
\draw (6,-7) + (0.5,0.5) node{\small D};
\draw[thin] (-4.75,-7) -- (8,-7);
\draw (-3.3,-8) + (0, 0.3) node{\small Prefix};
\draw (-3.3,-8) + (0, -0.3) node{\small beam search};
\draw (-1.5,-8) + (0.5,0.5) node{\small $A_B$};
\draw (3,-8) + (0.5,0.5) node{\small ?};
\draw (-1.5,-9) + (0.5,0.5) node{\small $Y_B$};
\draw (2,-9) + (0.5,0.5) node{\small E};
\draw (3,-9) + (0.5,0.5) node{\small E};
\draw (4,-9) + (0.5,0.5) node{\small D};
\draw[thin] (-4.75,-9) -- (8,-9);
\draw (-3.3,-10) + (0, 0.6) node{\small Extended};
\draw (-3.3,-10) + (0, 0) node{\small prefix};
\draw (-3.3,-10) + (0, -0.5) node{\small beam search};
\draw (-1.5,-10) + (0.5,0.5) node{\small $A_E$};
\draw (0,-10) + (0.5,0.5) node{\small E};
\draw (1,-10) + (0.5,0.5) node{\small E};
\draw (2,-10) + (0.5,0.5) node{\small $\epsilon$};
\draw (3,-10) + (0.5,0.5) node{\small E};
\draw (4,-10) + (0.5,0.5) node{\small $\epsilon$};
\draw (5,-10) + (0.5,0.5) node{\small D};
\draw (6,-10) + (0.5,0.5) node{\small D};
\draw (7,-10) + (0.5,0.5) node{\small D};
\draw (-1.5,-11) + (0.5,0.5) node{\small $Y_E$};
\draw (0.5,-11) + (0.5,0.5) node{\small E};
\draw (3,-11) + (0.5,0.5) node{\small E};
\draw (6,-11) + (0.5,0.5) node{\small D};
\end{tikzpicture}
\caption{An example with (1) dataset represented by data and label with corresponding alignment $A_L$ and collapsed token sequence $Y_L$, (2) the single stage approach for intake gesture detection with estimated probabilities $p(c|x_t)$, and alignments as well as collapsed token sequences produced by \textit{greedy decoding}, \textit{prefix beam search} as well as \textit{extended prefix beam search}. Note that finding the alignment $A_E$ produced by \textit{extended prefix beam search} is the key element missing for simple \textit{prefix beam search}.}
\label{fig:example}
\end{figure}

\subsection{Greedy decoding}

During inference, we decode the probabilities $p(c|x_t)$ into a sequence of tokens $Y$.
This can be interpreted as choosing an alignment $A$, which is then collapsed to $Y$.
A fast and simple solution is \textit{Greedy decoding}, which chooses the alignment by selecting the maximum probability token at each time step $t$ \cite{hannun2017sequence}.
However, this method is not guaranteed to produce the most probable $Y$, since it does not take into account that each $Y$ can have many possible alignments.
In the example of Fig \ref{fig:example}, greedy decoding gives the alignment $[E,E,\epsilon,\epsilon,\epsilon,D,D,D]$ which collapses to $[E,D]$.
Using Equation \ref{eq:prob}, we can compute that this is indeed an inferior solution to $[E,E,D]$.\footnote{Specifically, applying CTC loss to the numerical example in Fig. \ref{fig:example}, we find that $p([E,D]|X) \approx 0.0719 < 0.1305 \approx p([E,E,D]|X)$}

\subsection{Prefix beam search}

Traversing all possible alignments turns out to be infeasible due to their large number \cite{hannun2017sequence}.
The \textit{prefix beam search} algorithm \cite{graves2006connectionist} uses dynamic programming to search for a token sequence $\hat{Y}$ that maximises $p(\hat{Y}|X)$.
It presents a trade-off between computation and solution quality, which can be adjusted through the beam width $k$, determining how many possible solutions the algorithm remembers.
Prefix beam search with a beam width of 1 is equivalent to greedy decoding.
However, it is important to note that prefix beam search does not remember specific alignments.
Hence, it is not possible to temporally localize intake events (see missing $A_B$ in Fig. \ref{fig:example}).

The algorithm determines beams in terms of \textit{prefixes} $\ell$ (candidates for the output token sequence $\hat{Y}$ up to time $t$), which are stored in a list $Y$.
Each prefix is associated with two probabilities, the first of ending in a blank, $p_b(\ell|x_{1:t})$, and the second of not ending in a blank, $p_{nb}(\ell|x_{1:t})$.
For each time step $t$, the algorithm updates the probabilities for every prefix in $Y$ for the different cases of (i) adding a repeated token and (ii) adding a blank, and adds possible new prefixes.
Due to the algorithm design, branches with equal prefixes are dynamically merged.
The algorithm then keeps the $k$ best updated prefixes.

\subsection{Extended prefix beam search}

Standard prefix beam search finds a token sequence $\hat{Y}$, without retaining information about the alignments $A_{X,\hat{Y}}$.
In order to infer the \textit{timing} of the decoded events in a way consistent with CTC loss, the goal of our extended prefix beam search is to find $\hat{A}$.
This is the most probable alignment that could have produced $\hat{Y}$, as expressed by Equation \ref{eq:alignment}.

\vskip -.6\baselineskip plus -1fil

\begin{equation}
	\hat{A} = \argmax_{A_{X,\hat{Y}}} \prod_{t=1}^{T}p(c=a_t|x_t)
	\label{eq:alignment}
\end{equation}

\vskip -.4\baselineskip plus -1fil
 
Instead of running a separate algorithm based on $\hat{Y}$, we search for $\hat{A}$ simultaneously as part of prefix beam search, which already includes most of the necessary computation.
We add two additional lists for each beam $\ell$, $A_b(\ell)$ and $A_{nb}(\ell)$, which store alignment candidates that resolve to $\ell$ as well as their corresponding probabilities.
Every time a probability is updated in prefix beam search, we add new alignment candidates and associated probabilities to the appropriate lists.
This includes (i) adding a repeated token, (ii) adding a blank token, and (iii) adding a token that extends the prefix. 
The algorithm design implies that if two beams with identical prefixes are merged, alignment candidates are also merged dynamically.
At the end of each time step $t$, we resolve the alignment candidates for each $\ell$ in $Y$ by choosing the highest probability for each $A_b(\ell)$ and $A_{nb}(\ell)$.
Finally, for each of the $k$ best token sequences in $Y$, the best alignment candidate $\hat{A}$ is chosen as the more probable one out of $A_b(\ell)$ and $A_{nb}(\ell)$.

We created a Python implementation\footnote{See \url{https://gist.github.com/prouast/a73354a7586cc6bc444d2013001616b7}} of the pseudo-code shown in Algorithm 1.
Note that this version is not created with efficiency in mind.
For our experiments, we implemented a more efficient implementation\footnote{Available at \url{https://github.com/prouast/ctc-beam-search-op}} as a C++ TensorFlow kernel.

\subsection{Network architectures}

Although they are trained with different loss functions, both the single-stage and two-stage approaches each rely on an underlying deep neural network which estimates probabilities.
We choose adapted versions of the ResNet architecture \cite{he2016deep}.
Our video network is a CNN-LSTM with a ResNet-50 backbone adjusted for our video resolution.
For inertial data, we use a CNN-LSTM with a ResNet-10 backbone using 1D convolutions.
Table \ref{tab:architecture} reports the parameters and output sizes for all layers. 

\begin{table}[t]
\centering
\caption{Architectures for our single-stage and two-stage models}
\label{tab:architecture}
\begin{threeparttable}
\setlength{\extrarowheight}{3pt}
\newcommand{\ns}{\negthickspace}
\newcommand{\stimes}{{\mkern-1mu\times\mkern-1mu}}
\begin{tabular}{ p{.6cm} | p{1.35cm} p{1.65cm} | p{.7cm} p{1.05cm} p{1.05cm}}
 & \multicolumn{2}{c|}{Video} & \multicolumn{3}{c}{Inertial} \\
Layer & \multicolumn{2}{c|}{ResNet-50 CNN-LSTM} & \multicolumn{3}{c}{ResNet-10 CNN-LSTM} \\
& \multicolumn{2}{c|}{OREBA} & & OREBA & Clemson \\ 
& params & output size & params & \parbox[t]{1.3cm}{output size} & \parbox[t]{1.3cm}{output size} \\
\hline
data & & $16\times128^2\times3$ & & $512\times12$ & $120\times6$ \\
\hline
conv1 & \parbox[t]{1cm}{$5^2,64$\\stride $1^2$} & $16\times128^2\times64\ns$ & \parbox[t]{1cm}{$1,64$\\stride $1$} & $512\times64$ & $120\times64$ \\ [.3cm]
\hline
pool1 & \parbox[t]{1cm}{$2^2$\\stride $2^2$} & $16\times64^2\times64$ & & & \\ [.3cm]
\hline
conv2 & $\ns\left[\begin{matrix} 1^2, 64 \\ 3^2, 64 \\ 1^2, 256 \end{matrix}\right]\ns\stimes3$ & $16\times64^2\times256\ns$ & $\ns\left[\begin{matrix} 3, 64 \\ 3, 64\end{matrix}\right]$ & $512\times64\ns$ & $120\times64$ \\
\hline
conv3 & $\ns\left[\begin{matrix} 1^2, 128 \\ 3^2, 128 \\ 1^2, 512 \end{matrix}\right]\ns\stimes4$ & $16\times32^2\times512\ns$ & $\ns\left[\begin{matrix} 3, 128 \\ 3, 128\end{matrix}\right]$ & $256\times128\ns$ & $120\times128\ns$ \\
\hline
conv4 & $\ns\left[\begin{matrix} 1^2, 256 \\ 3^2, 256 \\ 1^2, 1024 \end{matrix}\right]\ns\stimes6$ & $16\stimes16^2\stimes1024\ns$ & $\ns\left[\begin{matrix} 5, 256 \\ 5, 256\end{matrix}\right]$ & $128\times256\ns$ & $60\times256$ \\
\hline
conv5 & $\ns\left[\begin{matrix} 1^2, 512 \\ 3^2, 512 \\ 1^2, 2048 \end{matrix}\right]\ns\stimes3$ & $16\times8^2\times2048\ns$ & $\ns\left[\begin{matrix} 5, 512 \\ 5, 512\end{matrix}\right]$ & $64\times512$ & $60\times512$ \\
\hline
\parbox[t]{.6cm}{pool} & & $16\times2048$ & & & \\
\hline
lstm & & $16\times128$ & & $64\times64$ & $60\times64$ \\
\hline
dense\tnote{a} & & $16\times |\Sigma|$ & & $64\times |\Sigma|$ & $60\times |\Sigma|$ \\
\end{tabular}
\begin{tablenotes}
\item[a] $\Sigma$ includes the blank token, hence $|\Sigma|=2$ for generic intake gesture detection and $|\Sigma|=3$ for detection of eating and drinking gestures.
\end{tablenotes}
\end{threeparttable}
\end{table}

\begin{algorithm*}[htp]
\label{alg:extctc}
\DontPrintSemicolon
\SetAlCapHSkip{0em}
\KwData{Probability distributions $p(c|x_t)$ for tokens $c \in \Sigma$ in sensor data $x_t$ from $t=1,\dots,T$.}
\KwResult{$k$ best decoded sequences of tokens $Y$ and best corresponding alignments $A$.}
 $p_b(\emptyset|x_{1:0})\leftarrow1$, $p_{nb}(\emptyset|x_{1:0})\leftarrow0$\;
 $Y\leftarrow\{\emptyset\}$\;
 $A_b(\emptyset)\leftarrow\{(\emptyset,1)\}$, $A_{nb}(\emptyset)\leftarrow\{(\emptyset,1)\}$\;
 \For{$t=1,\dots,T$}{
 	$Y'\leftarrow\{\}$\;
 	$A'_b(\cdot)\leftarrow\{\}$, $A'_{nb}(\cdot)\leftarrow\{\}$\;
 	\For{$\ell$ {\bf in} $Y$}{
 		\If{$\ell \notin Y'$}{
 			add $\ell$ to $Y'$\;
 		}
 		\If{$\ell\neq\emptyset$}{
 			$p_{nb}(\ell|x_{1:t})\leftarrow p_{nb}(\ell|x_{1:t})+p_{nb}(\ell|x_{1:t-1})p(\ell_{\vert\ell\vert}|x_{1:t})$\;
 			add $($ concatenate $A_{nb}(\ell)$ and $\ell_{\vert\ell\vert}$, $p(A_{nb}(\ell))p(\ell_{\vert\ell\vert}|x_{1:t})$ $)$ to $A'_{nb}(\ell)$\;
 		}
 		$p_b(\ell|x_{1:t})\leftarrow p_b(\ell|x_{1:t})+p(\epsilon|x_{1:t})(p_b(\ell|x_{1:t-1})+p_{nb}(\ell|x_{1:t-t}))$\;
 		add $($ concatenate $A_{b}(\ell)$ and $\epsilon$, $p(A_{b}(\ell))p(\epsilon|x_{1:t})$ $)$ to $A'_{b}(\ell)$\;
 		add $($ concatenate $A_{nb}(\ell)$ and $\epsilon$, $p(A_{nb}(\ell))p(\epsilon|x_{1:t})$ $)$ to $A'_{b}(\ell)$\;
 		\For{$c$ {\bf in} $\Sigma\setminus\epsilon$}{
 			$\ell^+\leftarrow$ concatenate $\ell$ and $c$\; 
 			add $\ell^+$ to $Y'$\;
 			\eIf{$\ell\neq\emptyset$ {\bf and} $c=\ell_{\vert\ell\vert}$}{
 				$p_{nb}(\ell^+|x_{1:t})\leftarrow p_{nb}(\ell^+|x_{1:t})+p_b(\ell|x_{1:t-1})p(c|x_{1:t})$\;
 				add $($ concatenate $A_{nb}(\ell)$ and $c$, $p(A_{b}(\ell))p(c|x_{1:t})$ $)$ to $A'_{nb}(\ell^+)$\;
 			}{
 				$p_{nb}(\ell^+|x_{1:t})\leftarrow p_{nb}(\ell^+|x_{1:t})+p(c|x_{1:t})(p_b(\ell|x_{1:t-1})+p_{nb}(\ell|x_{1:t-1}))$\;
 				add $($ concatenate $A_b(\ell)$ and $c$, $p(A_{b}(\ell))p(c|x_{1:t})$ $)$ to $A'_b(\ell^+)$\;
 				add $($ concatenate $A_{nb}(\ell)$ and $c$, $p(A_{nb}(\ell))p(c|x_{1:t})$ $)$ to $A'_{nb}(\ell^+)$\;
 			}
 		}
 	}
 	$Y\leftarrow k$ most probable prefixes in $Y'$\;
 	\For{$\ell$ {\bf in} $Y$}{
 		$A_b(\ell)\leftarrow$ the most probable sequence in $A'_b(\ell)$\;
 		$A_{nb}(\ell)\leftarrow$ the most probable sequence in $A'_{nb}(\ell)$\;
 	}
 }
\For{$\ell$ {\bf in} $Y$}{
 	$A(\ell)\leftarrow$ the most probable sequence in $\{A_b(\ell), A_{nb}(\ell)\}$\;
}
\Return $Y$, $A$\;
\parbox{\linewidth}{\caption{Extended prefix beam search algorithm (loosely based on \cite{hannun2014firstpass}): The algorithm stores current prefixes in $Y$. Probabilities are stored and updated in terms of prefixes ending in blank $p_b(\ell|x_t)$ and non-blank $p_{nb}(\ell|x_t)$, facilitating dynamic merging of beams with identical prefixes. The empty set is used to initialize $Y$ and associated with probability $1$ for blank, and $0$ for non-blank. $A_b(\ell)$ and $A_{nb}(\ell)$ store the current candidates for alignments (ending in blank and non-blank) pertaining to prefix $\ell$, along with their probabilities. They are likewise initialized for the empty prefix. The algorithm then loops over the time steps, updating the prefixes and associated alignments. Each current candidate $\ell$ is re-entered into the new prefixes $Y'$, adjusting the probabilities for repeated tokens and added blanks. The corresponding alignment candidates and their probabilities are added to the new alignment candidates $A'_{nb}(\ell)$ and $A'_b(\ell)$. Furthermore, for each non-blank token in $\Sigma$, a new prefix is created by concatenation, the probability is updated, and corresponding alignment candidates are added. At the end of each time step, we set $Y$ to the $k$ most probable prefixes in $Y'$ and resolve the alignment candidates for each of those prefixes as the most probable ones. Finally, for each of the $k$ best token sequences in $Y$, the best alignment candidate is chosen as the more probable one out of $A_b(\ell)$ and $A_{nb}(\ell)$.}}
\end{algorithm*}


\section{Experiments and analysis}
\label{sec:experiments}

In the experiments, we compare the proposed single-stage approach to the thresholding \cite{dong2012new} and the two-stage approach \cite{kyritsis2020modeling} \cite{heydarian2020deep} using two datasets of annotated intake gestures (OREBA \cite{rouast2019learning} and Clemson \cite{shen2017assessing}).
To this day, these are the largest publicly available datasets for intake gesture detection.
For both datasets, we attempt detection of generic intake gestures, as well as detection of eating and drinking gestures.
Across our experiments, we use time windows of \textit{8 seconds}, which ensures that examples regularly contain multiple intake events.
All code used for the experiments is available at \url{https://github.com/prouast/ctc-intake-detection}.

\subsection{Approaches}

\subsubsection{Thresholding approach}

We implemented the thresholding approach with four parameters as described by Dong et al. \cite{dong2012new} and Shen et al. \cite{shen2017assessing}, which only relies on angular velocity (wrist roll).
For each dataset, we used the training set to estimate the parameters $T_1$, $T_2$, $T_3$, and $T_4$.

\subsubsection{Two-stage approach}

SOTA results on OREBA \cite{rouast2019learning}\cite{heydarian2020deep} are based on 2 second time windows, which is not sufficient for the single-stage approach.
Hence, to facilitate a fair comparison, we also train several two-stage models based on 8 second time windows.
In particular, we use cross-entropy loss to train two-stage versions of our own architectures outlined in Table \ref{tab:architecture}, as well as the architectures proposed in Heydarian et al. \cite{heydarian2020deep}, Rouast et al. \cite{rouast2019learning}, and the adapted version of Kyritsis et al. \cite{kyritsis2020modeling} used in \cite{heydarian2020deep}.
Note that the latter was originally designed to be trained with additional sub-gesture labels which are not available for the Clemson and OREBA datasets.
These models are trained with cross-entropy loss.
Detections on the video level are reported according to the Stage 2 maximum search algorithm by \cite{kyritsis2020modeling}.
To facilitate multi-class comparison, we also extend the Stage 2 search by applying the same threshold to both intake gesture classes.

\subsubsection{Single-stage approach}

Our single-stage models are trained using CTC loss \cite{graves2006connectionist}.
One caveat of the single-stage approach is that it requires a longer time window than Stage 1 of the two-stage approach.
This is to ensure that multiple gestures regularly appear in the training examples, providing a signal for learning temporal relations.
At the same time, due to memory restrictions for the video model, longer time windows come with the drawback of having to reduce the sampling rate of the input data.
In light of this tradeoff, we considered different configurations and ultimately decided for a window size of 8 seconds.
\footnote{A window size of 8 seconds allows a video sampling rate of 2 fps and translates into a 74.7\% chance of seeing at least one example with multiple gestures per batch during video model training on OREBA. Details on the considered window sizes can be found in the Supplemental Material S1.}
For inference, the probabilities estimated for each temporal segment are decoded into an alignment using the \textit{Extended prefix beam search}, and then collapsed to yield event detections.
Based on an analysis on the validation set (see Section \ref{sec:experiments:sub:beamwidth}), we used a beam width of 3.
On the video level, we first aggregate detections from the individual alignments of sliding windows using frame-wise majority voting before collapse.

\subsection{Training and evaluation metrics}

\subsubsection{Training}

All networks are trained using the \textit{Adam} optimizer on the respective training set with batch size 128 for inertial and 16 for video.
We use an exponentially decreasing learning rate starting at 1e-3, except for the SOTA implementations where we use the learning rate settings reported by the authors \cite{heydarian2020deep}\cite{kyritsis2020modeling}\cite{rouast2019learning}.
We also use minibatch loss scaling, analogously to \cite{rouast2019learning}.
Hyperparameter and model selection is based on the validation set.

\subsubsection{Evaluation}

\begin{figure}[t]
\centering
\includegraphics[width=.8\columnwidth]{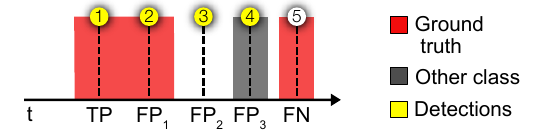}
\caption{The evaluation scheme (proposed by \cite{kyritsis2020modeling}; figure adapted from \cite{rouast2019learning}). (1) A true positive is the first detection within each ground truth event; (2) False positives of type 1 are further detections within the same ground truth event; (3) False positives of type 2 are detections outside ground truth events; (4) False positives of type 3 are detections made for the wrong class if applicable; (5) False negatives are non-detected ground truth events.}
\label{fig:scheme}
\end{figure}

\newcommand\tp{\mathit{TP}}
\newcommand\fn{\mathit{FN}}
\newcommand\fp{\mathit{FP}}

For comparison we use the $F_1$ measure, applying an extension of the evaluation scheme by Kyritsis et al. \cite{kyritsis2020modeling} (see Fig. \ref{fig:scheme}).
The scheme uses the ground truth to translate sparse detections into measurable metrics for a given label class.
As Rouast and Adam \cite{rouast2019learning} report, one correct detection per ground truth event counts as a true positive ($\tp$), while further detections within the same ground truth event are false positives of type 1 ($\fp_1$).
Detections outside ground truth events are false positives of type 2 ($\fp_2$) and non-detected ground truth events count as false negatives ($\fn$).
We extended the original scheme to support the multi-class case, where detections of a wrong class are false positives of type 3 ($\fp_3$).
Using the aggregate counts, we calculate precision, recall, and $F_1$.

\begin{table*}[h!]
\centering
\caption{Results for the OREBA and CLEMSON datasets (test set)}
\label{tab:results}
\begin{threeparttable}
\begin{tabular}{ l | c | c || c || c | c | c }
& & & Intake gestures & \multicolumn{3}{c}{(E)ating and (D)rinking gestures} \\
Method & Dataset & Modality & $F_1$ & $F_1^E$ & $F_1^D$ & $F_1^{E \land D}$ \\
\hline
Thresholding \cite{dong2012new} ($T_1=25$, $T_2=-25$, $T_3=2$, $T_4=2$, 64 Hz) & OREBA & Inertial & \calc{\fTOREBAIneThrGen}{3} & & & \\ 
Two-stage CNN-LSTM \cite{heydarian2020deep} (2 sec @ 64 Hz)\tnote{a} & OREBA & Inertial & \calc{\fTOREBAIneTwoSOTAGen}{3} & & & \\ 
Two-stage CNN-LSTM \cite{kyritsis2020modeling} (our implementation, 8 sec @ 64 Hz) & OREBA & Inertial & \calc{\fTOREBAIneKyrTwoGen}{3} & \calc{\fTOREBAIneKyrTwoEat}{3} & \calc{\fTOREBAIneKyrTwoDri}{3} & \calc{\fTOREBAIneKyrTwoEatDri}{3} \\ 
Two-stage CNN-LSTM \cite{heydarian2020deep} (our implementation, 8 sec @ 64 Hz) & OREBA & Inertial & \calc{\fTOREBAIneHeyTwoGen}{3} & \calc{\fTOREBAIneHeyTwoEat}{3} & \calc{\fTOREBAIneHeyTwoDri}{3} & \calc{\fTOREBAIneHeyTwoEatDri}{3} \\ 
Two-stage ResNet-10 CNN-LSTM (ours, 8 sec @ 64 Hz) & OREBA & Inertial & \calc{\fTOREBAIneTwoGen}{3} & \calc{\fTOREBAIneTwoEat}{3} & \calc{\fTOREBAIneTwoDri}{3} & \calc{\fTOREBAIneTwoEatDri}{3} \\ 
Single-stage ResNet-10 CNN-LSTM (ours, 8 sec @ 64 Hz) & OREBA & Inertial & \textbf{\calc{\fTOREBAIneSinGen}{3}} & \textbf{\calc{\fTOREBAIneSinEat}{3}} & \textbf{\calc{\fTOREBAIneSinDri}{3}} & \textbf{\calc{\fTOREBAIneSinEatDri}{3}} \\ 
\hline
Two-stage ResNet-50 SlowFast \cite{rouast2019learning} (2 sec @ 8 fps)\tnote{a} & OREBA & Video & \calc{\fTOREBAVidTwoSOTAGen}{3} & & & \\ 
Two-stage ResNet-50 SlowFast \cite{rouast2019learning} (our implementation, 8 sec @ 2 fps) & OREBA & Video & \calc{\fTOREBAVidSloTwoGen}{3} & \calc{\fTOREBAVidSloTwoEat}{3} & \calc{\fTOREBAVidSloTwoDri}{3} & \calc{\fTOREBAVidSloTwoEatDri}{3} \\ 
Two-stage ResNet-50 CNN-LSTM (ours, 8 sec @ 2 fps) & OREBA & Video & \calc{\fTOREBAVidTwoGen}{3} & \calc{\fTOREBAVidTwoEat}{3} & \textbf{\calc{\fTOREBAVidTwoDri}{3}} & \calc{\fTOREBAVidTwoEatDri}{3} \\ 
Single-stage ResNet-50 CNN-LSTM (ours, 8 sec @ 2 fps) & OREBA & Video & \textbf{\calc{\fTOREBAVidSinGen}{3}} & \textbf{\calc{\fTOREBAVidSinEat}{3}} & \calc{\fTOREBAVidSinDri}{3} & \textbf{\calc{\fTOREBAVidSinEatDri}{3}} \\ 
\hline
Thresholding \cite{dong2012new} ($T_1=15$, $T_2=-15$, $T_3=1$, $T_4=4$, 15 Hz) & Clemson & Inertial & \calc{\fTClemsonIneThrGen}{3} & & & \\ 
Two-stage CNN-LSTM \cite{kyritsis2020modeling} (our implementation, 8 sec @ 15 Hz) & Clemson & Inertial & \calc{\fTClemsonIneKyrTwoGen}{3} & \calc{\fTClemsonIneKyrTwoEat}{3} & \calc{\fTClemsonIneKyrTwoDri}{3} & \calc{\fTClemsonIneKyrTwoEatDri}{3} \\ 
Two-stage CNN-LSTM \cite{heydarian2020deep} (our implementation, 8 sec @ 15 Hz) & Clemson & Inertial & \calc{\fTClemsonIneHeyTwoGen}{3} & \calc{\fTClemsonIneHeyTwoEat}{3} & \calc{\fTClemsonIneHeyTwoDri}{3} & \calc{\fTClemsonIneHeyTwoEatDri}{3} \\ 
Two-stage ResNet-10 CNN-LSTM (ours, 8 sec @ 15 Hz) & Clemson & Inertial & \calc{\fTClemsonIneTwoGen}{3} & \calc{\fTClemsonIneTwoEat}{3} & \calc{\fTClemsonIneTwoDri}{3} & \calc{\fTClemsonIneTwoEatDri}{3} \\ 
Single-stage ResNet-10 CNN-LSTM (ours, 8 sec @ 15 Hz) & Clemson & Inertial & \textbf{\calc{\fTClemsonIneSinGen}{3}} & \textbf{\calc{\fTClemsonIneSinEat}{3}} & \textbf{\calc{\fTClemsonIneSinDri}{3}} & \textbf{\calc{\fTClemsonIneSinEatDri}{3}} \\ 
\hline
\end{tabular}
\begin{tablenotes}
\item[a] Test set results as reported in \cite{rouast2020oreba}. These models use time windows of 2 seconds, while single-stage models require 8 seconds due to their nature.
\end{tablenotes}
\end{threeparttable}
\end{table*}

\subsection{Datasets}

\subsubsection{OREBA}

The OREBA dataset \cite{rouast2020oreba} includes inertial and video data.
This dataset was approved by the IRB at The University of Newcastle on 10 September 2017 (H-2017-0208).
Specifically, we use the OREBA-DIS scenario with data for 100 participants (69 male, 31 female) and 4790 annotated intake gestures.
The split suggested by the dataset authors \cite{rouast2020oreba} includes training, validation, and test sets of 61, 20, and 19 participants.
For the inertial models, we use the processed\footnote{Processing includes mirroring for data uniformity, removal of the gravity effect using Madgwick's filter \cite{madgwick2010efficient}, and standardization.} accelerometer and gyroscope data from both wrists at 64 Hz (8 seconds correspond to 512 frames).
For the video models, we downsample the 140x140 pixel recordings from 24 fps to 2 fps (8 seconds correspond to 16 frames).
For data augmentation, we use random mirroring of the wrist for inertial data and the same steps as \cite{rouast2019learning} for video data, which includes spatial cropping to 128x128 pixels.

\subsubsection{Clemson}

The publicly available Clemson dataset \cite{shen2017assessing} consists of 488 annotated eating sessions across 264 participants (127 male, 137 female), a total of 20644 intake gestures.
Sensor data for accelerometer and gyroscope is available for the dominant hand at 15 Hz (8 seconds correspond to 120 frames).
We apply the same preprocessing and data augmentation as for OREBA.
We split the sessions into training, validation, and test sets (302, 93 and 93 sessions respectively) such that each participant appears in only one of the three (see Supplementary Material S3).
Note that because the Clemson dataset does not specify a dataset split, an alternative approach to test our models would have been k-fold cross-testing.
However, we decided for a specific split approach because (1) there is no data scarcity in the Clemson dataset that would require k-fold cross-testing, and (2) applying k-fold cross-testing on the Clemson dataset would be prohibitively expensive.
However, a shortcoming of this approach is that the results reported in Table \ref{tab:results} only reflect the test set which was selected by ourselves, not the original dataset authors.

\subsection{Results}

Results are listed in Table \ref{tab:results}, and extended results with detailed metrics are available in Supplementary Material S2.

\subsubsection{Detecting intake gestures}
\label{sec:experiments:sub:oreba-1}

The results for detecting only one generic intake event class are displayed in the center column of Table \ref{tab:results}.
We can see that the single stage approach generally yields higher performance than the thresholding and two stage approaches:
Relative improvements range between \relImpTOwnOREBAVidGen\% (\impDetTOwnOREBAVidGen) and \relImpTOwnClemsonIneGen\% (\impDetTOwnClemsonIneGen) over two-stage versions of our own architectures, and between \relImpTSOTAClemsonIneGen\% (\impDetTSOTAClemsonIneGen) and \relImpTSOTAOREBAVidGen\% (\impDetTSOTAOREBAVidGen) over our implementations of the SOTA.

For OREBA, we can additionally refer to previously published SOTA results based on 2 second windows.
Relative improvements over these results for the inertial \cite{heydarian2020deep} and video \cite{rouast2019learning} modalities equal \relImpTOREBAIneSOTA\% (\fTOREBAIneTwoSOTAGen$\,\to\,$\calc{\fTOREBAIneSinGen}{3}) and \relImpTOREBAVidSOTA\% (\fTOREBAVidTwoSOTAGen$\,\to\,$\calc{\fTOREBAVidSinGen}{3}), respectively.

\newcommand{\relImpTClemsonIneGen}{\calc{(\fTClemsonIneSinGen-\fTClemsonIneTwoGen)/\fTClemsonIneTwoGen*100}{1}}

For Clemson, we are not aware of any SOTA models other than the thresholding approach \cite{dong2012new} \cite{shen2017assessing}.
It is not surprising that both the two-stage and single-stage approach outperform the thresholding approach by a large margin.
Thresholding exclusively relies on one gyroscope channel, while the deep learning models build on a larger number of parameters.
Consistent with the OREBA results, we find that the single-stage approach yields a relative improvement of \relImpTClemsonIneGen\% (\calc{\fTClemsonIneTwoGen}{3}$\,\to\,$\calc{\fTClemsonIneSinGen}{3}) over the two-stage models on the Clemson dataset.
It is worth noting that the $F_1$ scores are generally lower for Clemson than for OREBA, indicating that it is more challenging for intake gesture detection.
However, this may be related to the lower sampling rate in Clemson and the fact that data for both wrists is available for OREBA, while only the dominant wrist is included in Clemson.

\subsubsection{Detecting eating and drinking gestures}
\label{sec:experiments:sub:oreba-2}

\newcommand{\incDiffEatDrinkTOREBA}{\calc{((\fTOREBAIneTwoGen-\fTOREBAIneTwoEatDri)/\fTOREBAIneTwoGen+(\fTOREBAIneSinGen-\fTOREBAIneSinEatDri)/\fTOREBAIneSinGen+(\fTOREBAVidTwoGen-\fTOREBAVidTwoEatDri)/\fTOREBAVidTwoGen+(\fTOREBAVidSinGen-\fTOREBAVidSinEatDri)/\fTOREBAVidSinGen+(\fTOREBAIneKyrTwoGen-\fTOREBAIneKyrTwoEatDri)/\fTOREBAIneKyrTwoGen+(\fTOREBAIneHeyTwoGen-\fTOREBAIneHeyTwoEatDri)/\fTOREBAIneHeyTwoGen+(\fTOREBAVidSloTwoGen-\fTOREBAVidSloTwoEatDri)/\fTOREBAVidSloTwoGen)/7*100}{1}}
\newcommand{\incDiffEatDrinkTClemson}{\calc{((\fTClemsonIneTwoGen-\fTClemsonIneTwoEatDri)/\fTClemsonIneTwoGen+(\fTClemsonIneSinGen-\fTClemsonIneSinEatDri)/\fTClemsonIneSinGen+(\fTClemsonIneKyrTwoGen-\fTClemsonIneKyrTwoEatDri)/\fTClemsonIneKyrTwoGen+(\fTClemsonIneHeyTwoGen-\fTClemsonIneHeyTwoEatDri)/\fTClemsonIneHeyTwoGen)/4*100}{1}}

This task consists of localization and simultaneous classification of intake gestures as either eating or drinking.
As there are no previously published results for this more fine-grained classification on either dataset, we rely on comparison between the separately trained single-stage and two-stage versions of our own models, as well as our implementations of the SOTA.
In the right hand side columns of Table \ref{tab:results}, we report separate $F_1$ scores for eating and drinking individually, as well as both together.

Three main observations emerge.
Firstly, the single-stage approach outperforms the two-stage approach to an even larger extent for this task: Relative improvements range from \relImpTOwnOREBAVidEatDri\% (\impDetTOwnOREBAVidEatDri) to \relImpTOwnOREBAIneEatDri\% (\impDetTOwnOREBAIneEatDri) over two-stage version of our own architectures, and from \relImpTSOTAOREBAIneEatDri\% (\impDetTSOTAOREBAIneEatDri) to \relImpTSOTAOREBAVidEatDri\% (\impDetTSOTAOREBAVidEatDri) over our implementations of SOTA architectures.
Secondly, the increased difficulty of this task compared to the generic detection task is noticeable in the difference between the $F_1$ and $F_1^{E \land D}$ scores, an average decrease of \incDiffEatDrinkTOREBA\% for OREBA and \incDiffEatDrinkTClemson\% for Clemson.
Thirdly, there are generally few misclassifications between eating and drinking.
As indicated by Table \ref{tab:averaged}, the frequency of false positives of type 2 is higher than the frequency of false positives of type 3 by almost two orders of magnitude. 

Overall, the video models achieve the best results on OREBA.
However, when focusing specifically on drinking detection, the two-stage video model achieves a better result.
This may be due to the low number of drinking gestures in the test set, which causes $F_1^D$ to randomly vary from $F_1^{E \land D}$ for multiple of the models in Table \ref{tab:results}.

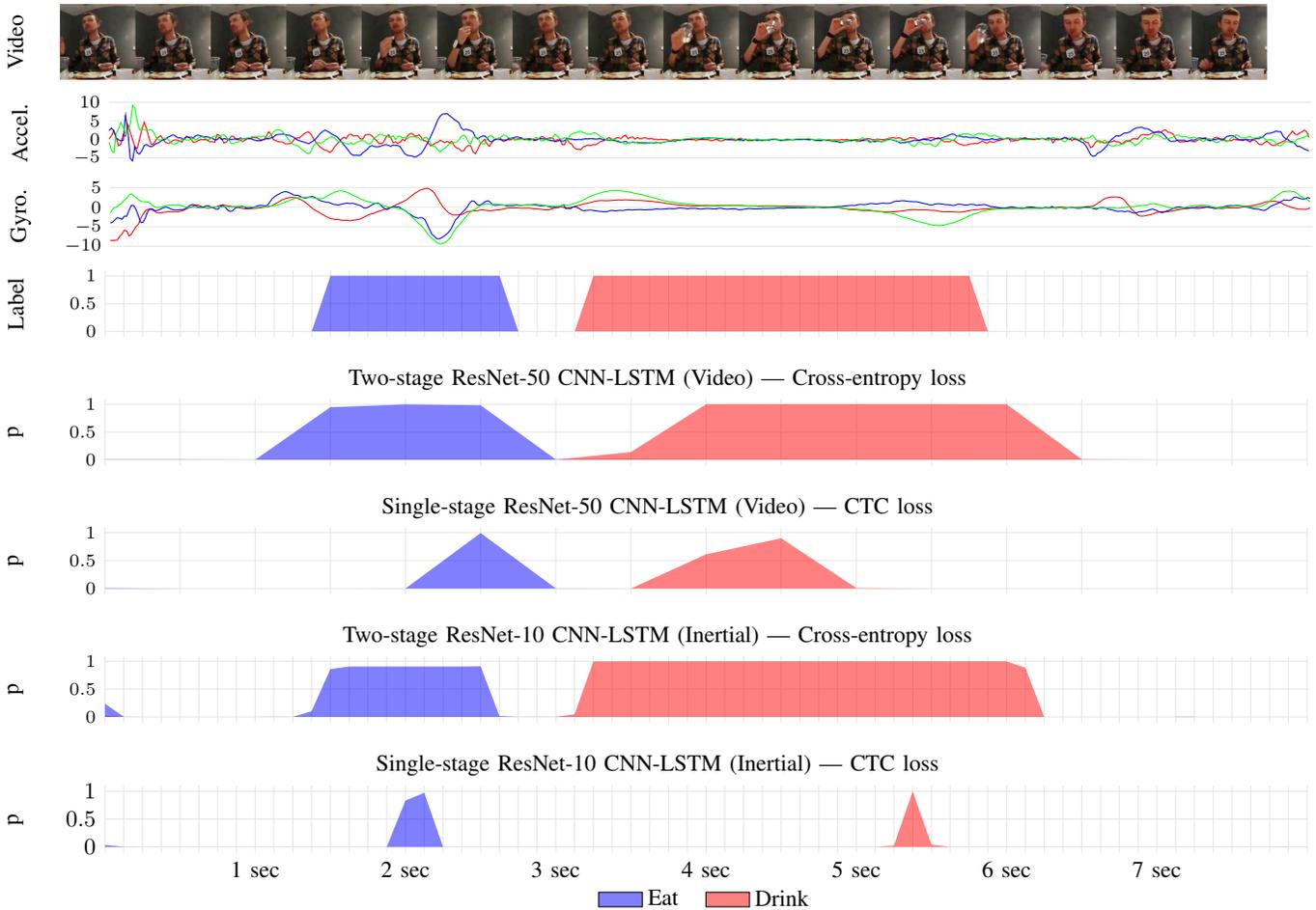
\begin{figure*}[ht]
\centering
\begin{tikzpicture}
\node (img)  {\includegraphics[width=\textwidth-.5cm]{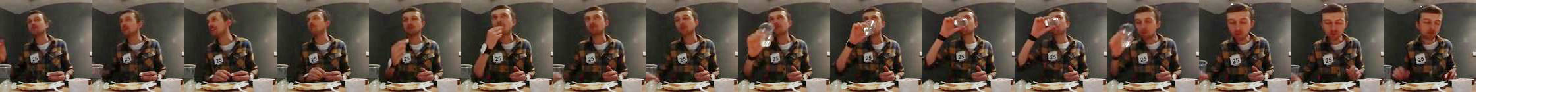}};
\node[left=of img, node distance=0cm, rotate=90, anchor=center,yshift=-0.5cm] {\small Video};
\end{tikzpicture}
\begin{tikzpicture}
\begin{axis}[ylabel=Accel., xmin=18944,xmax=19456,width=\textwidth, height=2.5cm, axis line style={draw=none}, xtick=\empty, ytick style={draw=none}, grid=both, major grid style={line width=.4pt,draw=black!10}, ticklabel style = {font=\scriptsize}, label style={font=\small}]
\addplot[color=red, mark=none] table[x=frame_id,y=dom_acc_x,col sep=comma]{figures/example/inertial_processed.csv};
\addplot[color=blue, mark=none] table[x=frame_id,y=dom_acc_y,col sep=comma]{figures/example/inertial_processed.csv};
\addplot[color=green, mark=none] table[x=frame_id,y=dom_acc_z,col sep=comma]{figures/example/inertial_processed.csv};
\end{axis}
\end{tikzpicture}
\begin{tikzpicture}
\begin{axis}[ylabel=Gyro., xmin=18944,xmax=19456,width=\textwidth, height=2.5cm, axis line style={draw=none}, xtick=\empty, ytick style={draw=none}, grid=both, major grid style={line width=.4pt,draw=black!10}, ticklabel style = {font=\scriptsize}, label style={font=\small}]
\addplot[color=red, mark=none] table[x=frame_id,y=dom_gyro_x,col sep=comma]{figures/example/inertial_processed.csv};
\addplot[color=blue, mark=none] table[x=frame_id,y=dom_gyro_y,col sep=comma]{figures/example/inertial_processed.csv};
\addplot[color=green, mark=none] table[x=frame_id,y=dom_gyro_z,col sep=comma]{figures/example/inertial_processed.csv};
\end{axis}
\end{tikzpicture}
\begin{tikzpicture}
\begin{axis}[ylabel=Label,xmin=1,xmax=65,width=\textwidth, height=2.5cm, axis line style={draw=none}, xtick={1,2,...,64,65}, xtick style={draw=none}, ytick style={draw=none}, xticklabels={}, grid=both, major grid style={line width=.4pt,draw=black!10}, ticklabel style = {font=\scriptsize}, label style={font=\small}]
\addplot[name path=eat,draw=none] table[x=frame,y=eat,col sep=comma]{figures/example/labels.csv};
\addplot[name path=drink,draw=none] table[x=frame,y=drink,col sep=comma]{figures/example/labels.csv};
\path[name path=zero] (axis cs:1,0) -- (axis cs:64,0);
\addplot[fill=blue,fill opacity=0.5] fill between[of=eat and zero];
\addplot[fill=red,fill opacity=0.5] fill between[of=drink and zero];
\end{axis}
\end{tikzpicture}
{\small Two-stage ResNet-50 CNN-LSTM (Video) --- Cross-entropy loss}
\begin{tikzpicture}
\begin{axis}[ylabel=p,xmin=1,xmax=17,width=\textwidth, height=2.5cm, axis line style={draw=none}, xtick style={draw=none}, ytick style={draw=none}, xticklabels={}, grid=both, major grid style={line width=.4pt,draw=black!10}, ticklabel style = {font=\scriptsize}, label style={font=\small}]
\addplot[name path=eat,draw=none] table[x=frame,y=eat,col sep=comma]{figures/example/logits_dgx_07.csv};
\addplot[name path=drink,draw=none] table[x=frame,y=drink,col sep=comma]{figures/example/logits_dgx_07.csv};
\path[name path=zero] (axis cs:1,0) -- (axis cs:16,0);
\addplot[fill=blue,fill opacity=0.5] fill between[of=eat and zero];
\addplot[fill=red,fill opacity=0.5] fill between[of=drink and zero];
\end{axis}
\end{tikzpicture}
{\small Single-stage ResNet-50 CNN-LSTM (Video) --- CTC loss}
\begin{tikzpicture}
\begin{axis}[ylabel=p,xmin=1,xmax=17,width=\textwidth, height=2.5cm, axis line style={draw=none}, xtick style={draw=none}, ytick style={draw=none}, xticklabels={}, grid=both, major grid style={line width=.4pt,draw=black!10}, ticklabel style = {font=\scriptsize}, label style={font=\small}]
\addplot[name path=eat,draw=none] table[x=frame,y=eat,col sep=comma]{figures/example/logits_dgx_05_02.csv};
\addplot[name path=drink,draw=none] table[x=frame,y=drink,col sep=comma]{figures/example/logits_dgx_05_02.csv};
\path[name path=zero] (axis cs:1,0) -- (axis cs:16,0);
\addplot[fill=blue,fill opacity=0.5] fill between[of=eat and zero];
\addplot[fill=red,fill opacity=0.5] fill between[of=drink and zero];
\end{axis}
\end{tikzpicture}
{\small Two-stage ResNet-10 CNN-LSTM (Inertial) --- Cross-entropy loss}
\begin{tikzpicture}
\begin{axis}[ylabel=p,xmin=1,xmax=65,width=\textwidth, height=2.5cm, axis line style={draw=none}, xtick={1,2,...,64,65}, xtick style={draw=none}, ytick style={draw=none}, xticklabels={}, grid=both, major grid style={line width=.4pt,draw=black!10}, ticklabel style = {font=\scriptsize}, label style={font=\small}]
\addplot[name path=eat,draw=none] table[x=frame,y=eat,col sep=comma]{figures/example/logits_hpc_21.csv};
\addplot[name path=drink,draw=none] table[x=frame,y=drink,col sep=comma]{figures/example/logits_hpc_21.csv};
\path[name path=zero] (axis cs:1,0) -- (axis cs:64,0);
\addplot[fill=blue,fill opacity=0.5] fill between[of=eat and zero];
\addplot[fill=red,fill opacity=0.5] fill between[of=drink and zero];
\end{axis}
\end{tikzpicture}
{\small Single-stage ResNet-10 CNN-LSTM (Inertial) --- CTC loss}
\begin{tikzpicture}
\begin{axis}[ylabel=p,xmin=1,xmax=65,width=\textwidth, height=2.5cm, axis line style={draw=none}, xtick={1,2,...,64,65}, xtick style={draw=none}, ytick style={draw=none}, xticklabels={,,,,,,,,1 sec,,,,,,,,2 sec,,,,,,,,3 sec,,,,,,,,4 sec,,,,,,,,5 sec,,,,,,,,6 sec,,,,,,,,7 sec}, grid=both, major grid style={line width=.4pt,draw=black!10}, ticklabel style = {font=\scriptsize}, label style={font=\small}, legend style={at={(0.5,-0.4)}, anchor=north,legend columns=-1, /tikz/every even column/.append style={column sep=0.3cm}, font=\small, legend style={draw=none}}, ticklabel style = {font=\small},]
\addplot[name path=eat,draw=none] table[x=frame,y=eat,col sep=comma]{figures/example/logits_hpc_13.csv};
\addplot[name path=drink,draw=none] table[x=frame,y=drink,col sep=comma]{figures/example/logits_hpc_13.csv};
\addplot[fill=blue,fill opacity=0.5] fill between[of=eat and zero];
\addplot[fill=red,fill opacity=0.5] fill between[of=drink and zero];
\path[name path=zero] (axis cs:1,0) -- (axis cs:64,0);
\legend{,,Eat,Drink}
\end{axis}
\end{tikzpicture}
\caption{Illustrating the effect of training with CTC loss or cross-entropy loss using input data, label, and model predictions for one 8 second example from the OREBA validation set.}
\label{fig:ctcvscrossent}
\end{figure*}

\newcommand{\avgFTTwo}{\calc{(\fTOREBAIneTwoGen+\fTOREBAIneTwoEatDri+\fTOREBAVidTwoGen+\fTOREBAVidTwoEatDri+\fTClemsonIneTwoGen+\fTClemsonIneTwoEatDri)/6.0}{4}}
\newcommand{\avgFTSinExt}{\calc{(\fTOREBAIneSinGen+\fTOREBAIneSinEatDri+\fTOREBAVidSinGen+\fTOREBAVidSinEatDri+\fTClemsonIneSinGen+\fTClemsonIneSinEatDri)/6.0}{4}}

\begin{table*}[h!]
\centering
\caption{Averaged results across all experiments (test set). Number of $TP$, $FP_1$, $FP_2$, $FP_3$, and $FN$ are expressed as percentages of the respective ground truth number of gestures to facilitate comparisons.}
\label{tab:averaged}
\begin{tabular}{ l | c | c | c | c | c | c }
Method & $TP$ [\%] & $FP_1$ [\%] & $FP_2$ [\%] & $FP_3$ [\%] & $FN$ [\%] & $F_1$ \\
\hline
Two-stage & 76.39 & 2.15 & 10.80 & 0.17 & 23.61 & \avgFTTwo \\ 
Single-stage, greedy decoding & 79.48 & \textbf{0.48} & \textbf{10.53} & \textbf{0.15} & 20.52 & \calc{\avgFTSinGre}{4} \\ 
Single-stage, extended prefix beam search & \textbf{80.58} & 0.49 & 11.76 & 0.15 & \textbf{19.42} & \textbf{\avgFTSinExt} \\ 
\hline
\end{tabular}
\end{table*}

\subsection{Effect of training with CTC loss or cross-entropy loss}
\label{sec:experiments:sub:effect}

During our introduction of CTC loss in Section \ref{sec:method:sub:ctc}, we mentioned that weakly supervised training with CTC causes our networks to learn a different approach of detecting events than cross-entropy loss.
We can think of cross-entropy loss as causing the network to predict \textit{whether a frame occurs anytime during} the gesture that is being detected.
The analogous way of thinking about CTC loss is to predict \textit{which frames are the most distinctive about} the gesture that is being detected.
This causes the signature for predictions by our single-stage models to look more like probability spikes, while the two-stage models produce sequences of high probability values.

We illustrate this characteristic difference between the single-stage and two-stage approaches in Fig. \ref{fig:ctcvscrossent} using an example from the validation set of OREBA for eating and drinking detection.
Here, time-synchronized 2 fps video and 64 Hz inertial data (dominant hand) for one 8 second time window are plotted alongside the ground truth and predictions of the corresponding two-stage and single-stage models.
Note that the output frequencies of the models differ, with 2 Hz for the video models and 8 Hz for the inertial models. 
We observe that the predictions by the two-stage models indeed mimic the ground truth, while the single-stage models produce probability spikes.
Furthermore, these probability spikes line up temporally with the patterns that appear to be most distinct about the gestures for the human eye.

\begin{figure}
  \begin{subfigure}[c]{.55\columnwidth}
 	\begin{tikzpicture}
	\begin{axis}[grid=both,
	  width=5.2cm, height=4cm,
  	  grid style={line width=.2pt, draw=black!10},
 	  major grid style={line width=.4pt,draw=black!10},
  	  axis lines=middle,
  	  compat=1.3,
 	  xtick={16,32,48}, xticklabels={},
  	  ytick={0.001, 0.5, 1.0}, yticklabels={0, 0.5, 1},
  	  xlabel=$t$, xlabel style={at={(ticklabel cs: 0.5)}, yshift=8pt, anchor=north},
  	  ylabel=$p$,
  	  enlargelimits={abs=0.1},
  	  ticklabel style = {font=\small},
  	  label style={font=\small}]
	\addplot[color=ctc_eat, mark=none, dashed] table[x=frame,y=mean,col sep=comma]{figures/aggregated/oreba-inert/hpc_13_01_model_best_37000_valid_1.csv};
	\addplot[color=ctc_drink, mark=none, dashed] table[x=frame,y=mean,col sep=comma]{figures/aggregated/oreba-inert/hpc_13_01_model_best_37000_valid_2.csv};
	\addplot[name path=ctc_eat_upper,draw=none] table[x=frame,y=q_75,col sep=comma]{figures/aggregated/oreba-inert/hpc_13_01_model_best_37000_valid_1.csv};
	\addplot[name path=ctc_eat_lower,draw=none] table[x=frame,y=q_25,col sep=comma]{figures/aggregated/oreba-inert/hpc_13_01_model_best_37000_valid_1.csv};
	\addplot[fill=ctc_eat,fill opacity=0.2] fill between[of=ctc_eat_upper and ctc_eat_lower];
	\addplot[name path=ctc_drink_upper,draw=none] table[x=frame,y=q_75,col sep=comma]{figures/aggregated/oreba-inert/hpc_13_01_model_best_37000_valid_2.csv};
	\addplot[name path=ctc_drink_lower,draw=none] table[x=frame,y=q_25,col sep=comma]{figures/aggregated/oreba-inert/hpc_13_01_model_best_37000_valid_2.csv};
	\addplot[fill=ctc_drink,fill opacity=0.2] fill between[of=ctc_drink_upper and ctc_drink_lower];
	\addplot[color=cross_eat, mark=none, dashed] table[x=frame,y=mean,col sep=comma]{figures/aggregated/oreba-inert/hpc_21_model_best_39000_valid_1.csv};
	\addplot[color=cross_drink, mark=none, dashed] table[x=frame,y=mean,col sep=comma]{figures/aggregated/oreba-inert/hpc_21_model_best_39000_valid_2.csv};
	\addplot[name path=cross_eat_upper,draw=none] table[x=frame,y=q_75,col sep=comma]{figures/aggregated/oreba-inert/hpc_21_model_best_39000_valid_1.csv};
	\addplot[name path=cross_eat_lower,draw=none] table[x=frame,y=q_25,col sep=comma]{figures/aggregated/oreba-inert/hpc_21_model_best_39000_valid_1.csv};
	\addplot[fill=cross_eat,fill opacity=0.2] fill between[of=cross_eat_upper and cross_eat_lower];
	\addplot[name path=cross_drink_upper,draw=none] table[x=frame,y=q_75,col sep=comma]{figures/aggregated/oreba-inert/hpc_21_model_best_39000_valid_2.csv};
	\addplot[name path=cross_drink_lower,draw=none] table[x=frame,y=q_25,col sep=comma]{figures/aggregated/oreba-inert/hpc_21_model_best_39000_valid_2.csv};
	\addplot[fill=cross_drink,fill opacity=0.2] fill between[of=cross_drink_upper and cross_drink_lower];
	\end{axis}
	\end{tikzpicture}
	\caption{Inertial (OREBA)}
    \vspace*{2mm}
  \end{subfigure}%
  \begin{subfigure}[c]{.45\columnwidth}
	\begin{tikzpicture}
	\begin{axis}[grid=both,
	  width=5.2cm, height=4cm,
  	  grid style={line width=.2pt, draw=black!10},
 	  major grid style={line width=.4pt,draw=black!10},
  	  axis lines=middle,
  	  compat=1.3,
 	  xtick={16,32,48}, xticklabels={},
  	  ytick={0.0, 0.5, 1.0}, yticklabels={},
  	  xlabel=$t$, xlabel style={at={(ticklabel cs: 0.5)}, yshift=8pt, anchor=north},
  	  enlargelimits={abs=0.1},
  	  ticklabel style = {font=\small},
  	  label style={font=\small}]
	\addplot[color=ctc_eat, mark=none, dashed] table[x=frame,y=mean,col sep=comma]{figures/aggregated/oreba-video/dgx_05_03_model_best_53000_valid_1.csv};
	\addplot[color=ctc_drink, mark=none, dashed] table[x=frame,y=mean,col sep=comma]{figures/aggregated/oreba-video/dgx_05_03_model_best_53000_valid_2.csv};
	\addplot[name path=ctc_eat_upper,draw=none] table[x=frame,y=q_75,col sep=comma]{figures/aggregated/oreba-video/dgx_05_03_model_best_53000_valid_1.csv};
	\addplot[name path=ctc_eat_lower,draw=none] table[x=frame,y=q_25,col sep=comma]{figures/aggregated/oreba-video/dgx_05_03_model_best_53000_valid_1.csv};
	\addplot[fill=ctc_eat,fill opacity=0.2] fill between[of=ctc_eat_upper and ctc_eat_lower];
	\addplot[name path=ctc_drink_upper,draw=none] table[x=frame,y=q_75,col sep=comma]{figures/aggregated/oreba-video/dgx_05_03_model_best_53000_valid_2.csv};
	\addplot[name path=ctc_drink_lower,draw=none] table[x=frame,y=q_25,col sep=comma]{figures/aggregated/oreba-video/dgx_05_03_model_best_53000_valid_2.csv};
	\addplot[fill=ctc_drink,fill opacity=0.2] fill between[of=ctc_drink_upper and ctc_drink_lower];
	\addplot[color=cross_eat, mark=none, dashed] table[x=frame,y=mean,col sep=comma]{figures/aggregated/oreba-video/dgx_07_model_best_72000_valid_1.csv};
	\addplot[color=cross_drink, mark=none, dashed] table[x=frame,y=mean,col sep=comma]{figures/aggregated/oreba-video/dgx_07_model_best_72000_valid_2.csv};
	\addplot[name path=cross_eat_upper,draw=none] table[x=frame,y=q_75,col sep=comma]{figures/aggregated/oreba-video/dgx_07_model_best_72000_valid_1.csv};
	\addplot[name path=cross_eat_lower,draw=none] table[x=frame,y=q_25,col sep=comma]{figures/aggregated/oreba-video/dgx_07_model_best_72000_valid_1.csv};
	\addplot[fill=cross_eat,fill opacity=0.2] fill between[of=cross_eat_upper and cross_eat_lower];
	\addplot[name path=cross_drink_upper,draw=none] table[x=frame,y=q_75,col sep=comma]{figures/aggregated/oreba-video/dgx_07_model_best_72000_valid_2.csv};
	\addplot[name path=cross_drink_lower,draw=none] table[x=frame,y=q_25,col sep=comma]{figures/aggregated/oreba-video/dgx_07_model_best_72000_valid_2.csv};
	\addplot[fill=cross_drink,fill opacity=0.2] fill between[of=cross_drink_upper and cross_drink_lower];
	\end{axis}
	\end{tikzpicture}
	\caption{Video (OREBA)}
	\vspace*{2mm}
  \end{subfigure}
  ~
  \begin{subfigure}[l]{.55\columnwidth}
    \begin{tikzpicture}
	\begin{axis}[grid=both,
	  width=5.2cm, height=4cm,
  	  grid style={line width=.2pt, draw=black!10},
 	  major grid style={line width=.4pt,draw=black!10},
  	  axis lines=middle,
  	  compat=1.3,
 	  xtick={16,32,48}, xticklabels={},
  	  ytick={0.001, 0.5, 1.0}, yticklabels={0, 0.5, 1},
  	  xlabel=$t$, xlabel style={at={(ticklabel cs: 0.5)}, yshift=8pt, anchor=north},
  	  ylabel=$p$,
  	  enlargelimits={abs=0.1},
  	  legend style={at={(1.6,0.9)}, anchor=north, /tikz/every even column/.append style={column sep=0.3cm}, font=\small, legend style={draw=none}, cells={align=center}},
  	  ticklabel style = {font=\small},
  	  label style={font=\small}]
	\addplot[color=ctc_eat, mark=none, dashed] table[x=frame,y=mean,col sep=comma]{figures/aggregated/clemson-inert/hpc_14_07_model_best_165000_valid_1.csv};
	\addplot[color=ctc_drink, mark=none, dashed] table[x=frame,y=mean,col sep=comma]{figures/aggregated/clemson-inert/hpc_14_07_model_best_165000_valid_2.csv};
	\addplot[name path=ctc_eat_upper,draw=none,forget plot] table[x=frame,y=q_75,col sep=comma]{figures/aggregated/clemson-inert/hpc_14_07_model_best_165000_valid_1.csv};
	\addplot[name path=ctc_eat_lower,draw=none,forget plot] table[x=frame,y=q_25,col sep=comma]{figures/aggregated/clemson-inert/hpc_14_07_model_best_165000_valid_1.csv};
	\addplot[fill=ctc_eat,fill opacity=0.2,forget plot] fill between[of=ctc_eat_upper and ctc_eat_lower];
	\addplot[name path=ctc_drink_upper,draw=none,forget plot] table[x=frame,y=q_75,col sep=comma]{figures/aggregated/clemson-inert/hpc_14_07_model_best_165000_valid_2.csv};
	\addplot[name path=ctc_drink_lower,draw=none,forget plot] table[x=frame,y=q_25,col sep=comma]{figures/aggregated/clemson-inert/hpc_14_07_model_best_165000_valid_2.csv};
	\addplot[fill=ctc_drink,fill opacity=0.2,forget plot] fill between[of=ctc_drink_upper and ctc_drink_lower];
	\addplot[color=cross_eat, mark=none, dashed] table[x=frame,y=mean,col sep=comma]{figures/aggregated/clemson-inert/hpc_23_model_best_35000_valid_1.csv};
	\addplot[color=cross_drink, mark=none, dashed] table[x=frame,y=mean,col sep=comma]{figures/aggregated/clemson-inert/hpc_23_model_best_35000_valid_2.csv};
	\addplot[name path=cross_eat_upper,draw=none,forget plot] table[x=frame,y=q_75,col sep=comma]{figures/aggregated/clemson-inert/hpc_23_model_best_35000_valid_1.csv};
	\addplot[name path=cross_eat_lower,draw=none,forget plot] table[x=frame,y=q_25,col sep=comma]{figures/aggregated/clemson-inert/hpc_23_model_best_35000_valid_1.csv};
	\addplot[fill=cross_eat,fill opacity=0.2,forget plot] fill between[of=cross_eat_upper and cross_eat_lower];
	\addplot[name path=cross_drink_upper,draw=none,forget plot] table[x=frame,y=q_75,col sep=comma]{figures/aggregated/clemson-inert/hpc_23_model_best_35000_valid_2.csv};
	\addplot[name path=cross_drink_lower,draw=none,forget plot] table[x=frame,y=q_25,col sep=comma]{figures/aggregated/clemson-inert/hpc_23_model_best_35000_valid_2.csv};
	\addplot[fill=cross_drink,fill opacity=0.2,forget plot] fill between[of=cross_drink_upper and cross_drink_lower];
	\legend{Eat (Single-stage),Drink (Single-stage),Eat (Two-stage),Drink (Two-stage)}
	\end{axis}
	\end{tikzpicture}
	\caption{Inertial (Clemson)}
  \end{subfigure}
  \caption{Aggregating the predicted probabilities within all eating and drinking events in the validation sets of OREBA and Clemson. Probabilities are aligned in time and linearly interpolated, based on which we plot the mean and $[q_{25},q_{75}]$ interval. The characteristic peaks for single-stage models trained on inertial data appear to be clustered in the second half of ground truth events, while they mainly fall in the first half for models trained on video data.}
  \label{fig:aggregated}
\end{figure}
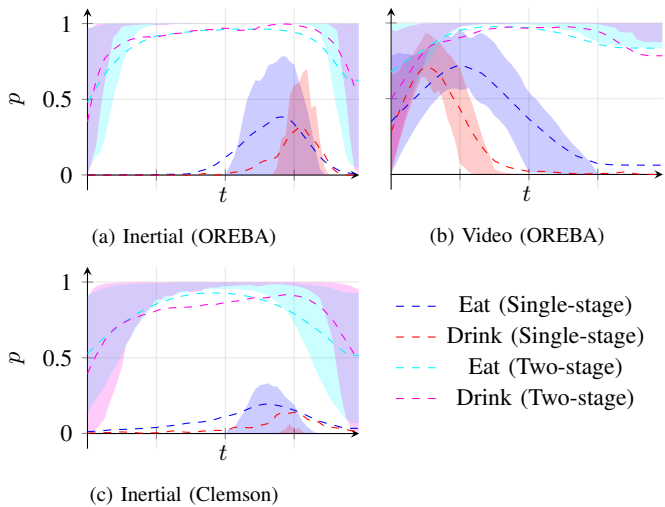

For a broader view of these characteristic differences between the single-stage and two-stage models, we use linear interpolation to aggregate the probabilities within all true positives in the validation set on a unitless timescale.
The distributions displayed in Fig. \ref{fig:aggregated} confirm that the two-stage models mimic the ground truth, while the probability spikes for single-stage models seem to be clustered in regions specific to the sensor modality.
While the probability spikes for the video models tend to fall in the first half of the ground truth events, those for the inertial models appear mainly in the second half.
This lends itself to the interpretation that video models target the frame in which ingestion takes place or the mouth is open (relatively early in the ground truth event), while inertial models leverage the characteristic downwards motion when finishing the intake gesture (relatively late).

When averaging the results across all datasets and tasks as reported in Table \ref{tab:averaged}, it becomes clear that training with CTC loss accounts for the majority of the improvement of single-stage models over two-stage models.
The effect of training with CTC loss manifests itself in a higher true positive rate and an associated lower false negative rate.
Furthermore, there is a significant drop in false positives of type 1, which were previously conjectured to be a restriction of the two-stage approach \cite{rouast2019learning}.
In particular, the single-stage approach avoids the predefined 2 second gap in Stage 2 of the two-stage approach and is thus less likely to lead to false positives of type 1 for gestures with a long duration.

\subsection{Difference between Greedy decoding and Extended beam search decoding}
\label{sec:experiments:sub:beamwidth}

\newcommand{\impAvgFVGreExt}{\calc{((\fVOREBAIneSinGen+\fVOREBAIneSinEatDri+\fVOREBAVidSinGen+\fVOREBAVidSinEatDri+\fVClemsonIneSinGen+\fVClemsonIneSinEatDri)/6.0-\avgFVSinGre)/\avgFVSinGre*100}{2}}

\begin{figure}
\centering
\begin{tikzpicture}
\begin{axis}[grid=both,
  grid style={line width=.2pt, draw=black!10},
  major grid style={line width=.4pt,draw=black!10},
  axis lines=middle,
  xtick={1,2,3,5,10,20},
  xticklabels={1,2,3,5,10,20},
  xlabel=Beam width,
  ylabel=Relative $F_1$ change,
  xmin=0,
  compat=1.3,
  ytick={0,.1,.2,.3,.4,.5},
  yticklabels={0.0\%,0.1\%,,0.3\%,,0.5\%},
  enlargelimits={abs=0.1},
  legend style={at={(0.5,-0.3)}, anchor=north,legend columns=-1, /tikz/every even column/.append style={column sep=0.3cm}, font=\small, legend style={draw=none}},     ticklabel style = {font=\small},
  label style={font=\small},
  width=6cm, height=4cm,
  legend style={cells={align=center}}]
\addplot[color=red, mark=*] table[x=beam_width,y=generic,col sep=comma]{figures/beam_width_data.csv};
\addplot[color=blue, mark=*] table[x=beam_width,y=eatdrink,col sep=comma]{figures/beam_width_data.csv};
\addplot[name path=generic_upper,draw=none] table[x=beam_width,y expr=\thisrow{generic}+\thisrow{generic_error}/2.0,col sep=comma]{figures/beam_width_data.csv};
\addplot[name path=generic_lower,draw=none] table[x=beam_width,y expr=\thisrow{generic}-\thisrow{generic_error}/2.0,col sep=comma]{figures/beam_width_data.csv};
\addplot[fill=red,fill opacity=0.2] fill between[of=generic_upper and generic_lower];
\addplot[name path=eatdrink_upper,draw=none] table[x=beam_width,y expr=\thisrow{eatdrink}+\thisrow{eatdrink_error}/2.0,col sep=comma]{figures/beam_width_data.csv};
\addplot[name path=eatdrink_lower,draw=none] table[x=beam_width,y expr=\thisrow{eatdrink}-\thisrow{eatdrink_error}/2.0,col sep=comma]{figures/beam_width_data.csv};
\addplot[fill=blue,fill opacity=0.2] fill between[of=eatdrink_upper and eatdrink_lower];
\legend{Generic intake gestures\\$|\Sigma|=2$,Eat and drink gestures\\$|\Sigma|=3$}
\end{axis}
\end{tikzpicture}
\caption{Average relative $F_1$ change with standard deviation when choosing different beam widths for decoding our models on the validation set. The base scenario is a beam width of 1, which corresponds to \textit{greedy decoding}. We observe that extended prefix beam search decoding mainly benefits the models for eating and drinking detection, and that there are no improvements for beam widths greater than 3.}
\label{fig:beamwidth}
\end{figure}
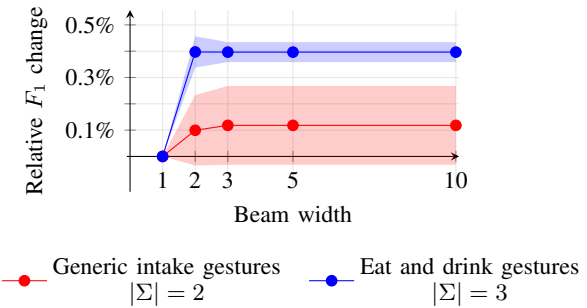

Recall that greedy decoding only considers the maximum probability token at each time step, which is equal to extended prefix beam search decoding with a beam width of 1.
As we increase the beam width, the algorithm considers more possible alignments and combines their probabilities if they lead to the same output sequence.
In theory, this means that the results produced by the extended prefix beam search decoding with a higher beam width better reflect the network's intended output than greedy decoding, since they are computed in the same way as CTC loss works internally.

To analyze the effect of different beam widths on the $F_1$ score and determine a beam width to use in our experiments, we decode our trained networks with different beam widths on the validation set.
As illustrated in Fig. \ref{fig:beamwidth}, the effect of extended prefix beam search decoding is not very noticeable - a relative improvement of only \impAvgFVGreExt\% on average.
In fact, there is no improvement for beam widths over 3, and hence we chose beam width 3 for decoding on the test set.

An explanation for these numbers may lie in the \textit{few classes} (i.e., only one or two types of gestures to be detected) and the associated relatively \textit{low uncertainty} exhibited by our scenario (i.e., limited variety of foods and environments).
This is also indicated by the low rate of false positives of type 3 in Table \ref{tab:averaged} and the high prediction confidences in Fig. \ref{fig:ctcvscrossent}.
It is well known that greedy decoding can work well as a heuristic in cases where most of the probability mass is allotted to a single alignment \cite{hannun2017sequence}.
It is evident from Fig. \ref{fig:beamwidth} that higher beam widths mainly benefited our task on eating and drinking gestures, which has one extra class and hence inherently carries more uncertainty.
Following this line of thought, it seems likely that the extended prefix beam search algorithm could lead to higher benefits over greedy decoding for datasets with more diverse labels and scenarios.


\section{Discussion}
\label{sec:discussion}

It is important to note that even though our implementations of the single-stage approach exhibit performance improvements compared to the two-stage approach, there are also several other differences between the two approaches that need to be considered in their application in research and practice.

First, the single-stage approach does not require detailed labels for the start and end timestamp of an intake gesture, but only a label for its apex.
These simplified labels can assist in reducing the effort in labelling new datasets or applying the approach in contexts where there are constraints on the sampling rate of the ground truth label (e.g. time-lapse recordings in field settings).

Second, while the probabilities provided by the two-stage approach align closely with the entire duration of the intake gesture as provided by the ground truth label, the single-stage approach only yields individual spikes within the intake gesture (see Fig. \ref{fig:ctcvscrossent}).
As such, the information provided by two-stage models is \textit{richer} in the sense that they allow to estimate the duration of the gesture as well as the timing between gestures, which is not possible with the single-stage approach.
For instance, if the spike in one gesture is towards the start of the ground truth event, and the spike in the subsequent gesture is towards the end, one would overestimate the gap between these gestures.
In other words, the simplified labels of the single-stage approach come with the caveat of simplified information in its predictions.

Third, both approaches rely on specific yet different assumptions related to the duration of eating gestures.
While the two-stage approach relies on a predefined gap between intake events (e.g., 2 seconds in \cite{kyritsis2020modeling} \cite{rouast2019learning}), the single-stage approach requires a window that is sufficiently large to likely capture a sequence of at least two intake gestures (e.g., 8 seconds).
The predefined gap of the two-stage approach creates the potential of inadvertently rejecting local probability maxima that are too close to each other.
By contrast, the large window of the single-stage approach comes with the drawback of increased memory requirements which are also reflected in the choice of 2 fps for the video models.


\section{Conclusion}
\label{sec:conclusion}

In this paper, we introduced a single-stage approach to detect intake gestures.
This is achieved by weakly supervised training of a deep neural network with CTC loss and decoding using a novel extended prefix beam search decoding algorithm.
Using CTC loss instead of cross-entropy loss allows us to interpret intake gesture detection as a sequence labelling problem, where the network labels an entire sequence as opposed to doing this independently in a frame-by-frame fashion.
Additionally, we are the first to attempt simultaneous detection of intake gestures and distinction between eating and drinking using deep learning.
We demonstrate improvements over the established two-stage approach \cite{kyritsis2020modeling} \cite{rouast2019learning} using two datasets.
These improvements apply to both generic intake gesture detection and eating/drinking detection tasks, and also to both video and inertial sensor data.

The proposed extended prefix beam search decoding algorithm is the second novel element in this context besides CTC loss.
This algorithm allows us to decode the probability estimate provided by the deep neural network in a way that is consistent with the computation of CTC loss.
However, despite the theoretical benefits of this algorithm, our results show that training with CTC loss accounts for the lion's share of the improvements we see over the two-stage approach.
This could be explained by the low number of classes for the datasets and tasks considered here.
Greedy decoding can hence be seen as a fast baseline alternative.
It remains to be seen in future work whether extended prefix beam search decoding is more useful when working with a larger number of classes and higher associated uncertainty.

While we used the CNN-LSTM framework for our models, one could also consider alternative architectures.
Importantly, the network must be able to cover the temporal context -- this makes it difficult to directly combine CTC loss with convolution-only models such as SlowFast \cite{salazar2019self}.
While CTC loss is traditionally combined with RNNs for this reason, Transformers have more recently emerged as another feasible choice \cite{salazar2019self}.
Another topic to be explored in future research is the effect of choosing different window sizes on model training and performance.

This work also has several other implications for future research.
We have shown a feasible way of localizing intake gestures while simultaneously classifying them as eating or drinking.
Given larger video datasets with more different food types and associated labels, future research could explore more fine-grained classification of different foods and gestures.
The necessity of large datasets has been pointed out \cite{shen2018impact} and detailed food classes are in fact available for the Clemson dataset, but tentative experiments indicated that inertial sensor data may not be sufficiently expressive to yield satisfactory results for food detection.
Another implication directly has to do with the practical task of labelling future datasets.
When working with CTC loss, events do not need to be painstakingly labelled with a start and end timestamp.
Instead, it is sufficient to mark the apex of the gesture -- similar to how the single-stage approach makes detections -- which has the potential to significantly reduce the labelling workload and reduce ambiguity around determining the exact start and end times of intake gestures.

\section*{Acknowledgment}

We gratefully acknowledge the support by the Bill \& Melinda Gates Foundation [OPP1171389].
This work was additionally supported by an Australian Government Research Training (RTP) Scholarship.

\bibliographystyle{assets/IEEEtran}
\bibliography{paper}


\begin{IEEEbiography}[{\includegraphics[width=1in,height=1.25in,clip,keepaspectratio]{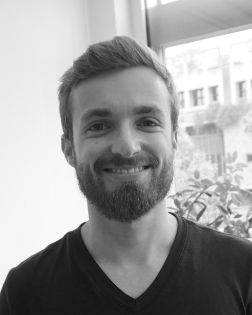}}]{Philipp V. Rouast}
received the B.Sc. and M.Sc. degrees in Industrial Engineering from Karlsruhe Institute of Technology, Germany, in 2013 and 2016 respectively.
He is currently working towards the Ph.D. degree in Information Systems and is a graduate research assistant at The University of Newcastle, Australia.
His research interests include deep learning, affective computing, HCI, and related applications of computer vision.
Find him at \url{https://www.rouast.com}.
\end{IEEEbiography}


\begin{IEEEbiography}[{\includegraphics[width=1in,height=1.25in,clip,keepaspectratio]{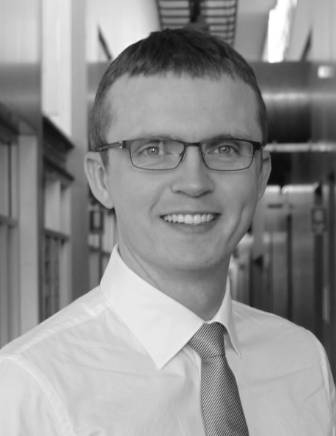}}]{Marc T. P. Adam}
received the undergraduate degree in computer science from the University of Applied Sciences W{\"u}rzburg, Germany, and the Ph.D. degree in information systems from the Karlsruhe Institute of Technology, Germany.
He is currently an Associate Professor of computing and information technology with The University of Newcastle, Australia.
His research interests include human-centered computing with applications in business, education, and health.
He is a Founding Member of the Society for NeuroIS.
\end{IEEEbiography}

\end{document}